\documentclass[11pt]{article}

\usepackage{booktabs}
\usepackage{multirow}
\usepackage{graphicx}
\usepackage[final]{acl}
\usepackage{amsthm}
\usepackage{algorithm}
\usepackage{algpseudocode}
\usepackage{amsmath, amssymb, amsfonts}
\usepackage{xcolor}
\theoremstyle{definition}  

\usepackage{enumitem}
\definecolor{darkgreen}{RGB}{34,139,34}
\usepackage{mathtools}
\definecolor{darkgreen}{RGB}{30, 120, 30} 
\definecolor{lightgray}{gray}{0.95}

\definecolor{oursblue}{RGB}{232, 245, 255}
\definecolor{oursblueheader}{RGB}{200, 230, 255}
\definecolor{rowgray}{RGB}{250, 250, 252}
\definecolor{sepline}{RGB}{210, 210, 210}
\definecolor{sectionline}{RGB}{170, 170, 170}
\definecolor{datasetgray}{RGB}{0, 0, 255}
\definecolor{impgreen}{RGB}{40, 167, 69}
\definecolor{avggreen}{RGB}{80, 160, 100}
\definecolor{avgrow}{RGB}{252, 252, 250}
\definecolor{goldstar}{RGB}{255, 193, 7}
\definecolor{headerblue}{RGB}{70, 130, 180}
\definecolor{improvgreen}{RGB}{40, 167, 69}
\definecolor{dropred}{RGB}{220, 53, 69}

\newcommand{\met}[2]{\textcolor{headerblue}{\textsf{#1}}\,$#2$}
\newcommand{\best}[1]{\textbf{#1}}
\newcommand{\imp}[1]{\textsuperscript{\textcolor{impgreen}{\scriptsize$\blacktriangle$\!#1}}}
\newcommand{\impavg}[1]{{\small\color{avggreen}$\blacktriangle$#1}}
\newcommand{\dataset}[1]{{\itshape\color{datasetgray}#1}}

\newcommand{\second}[1]{\underline{#1}}
\newcommand{\ours}{GeoEvolver}

\newcommand{\wo}[1]{\textit{w/o}~#1}

\newcommand{\drop}[1]{\textsuperscript{\textcolor{dropred}{\scriptsize$\blacktriangledown$\!#1}}}

\usepackage[most]{tcolorbox}
\usepackage{enumitem}
\newtcolorbox{promptbox}[2][]{
  colback=teal!5!white,
  colframe=teal!70!black,
  coltitle=white,
  title={#2},
  fonttitle=\bfseries,
  fontupper=\small\ttfamily,
  boxrule=0.8pt,
  rounded corners,
  arc=3pt,
  before skip=0.8em,
  after skip=0.8em,
  breakable,
  #1
}

\algrenewcommand\algorithmiccomment[1]{\hfill$\triangleright$ \textit{#1}}

\definecolor{phaseblue}{RGB}{0,102,204}

\usepackage{times}
\usepackage{latexsym}

\usepackage[T1]{fontenc}

\usepackage[utf8]{inputenc}

\usepackage{microtype}

\usepackage{inconsolata}

\usepackage{graphicx}
\usepackage[table]{xcolor} 
\title{Experience-Driven Multi-Agent Systems Are Training-free Context-aware Earth Observers}

\author{
\parbox{\linewidth}{\centering
  Pengyu Dai$^{1,2}$, Weihao Xuan$^{1,2}$, Junjue Wang$^{1}$, Hongruixuan Chen$^{1,2}$ \\
  Jian Song$^{2}$, Yafei Ou$^{2,3}$, Naoto Yokoya$^{1,2\dagger}$
}\\[0.5em]
\parbox{\linewidth}{\centering
  $^{1}$The University of Tokyo~
  $^{2}$RIKEN AIP ~
  $^{3}$Hokkaido University \\
  $^{\dagger}$Corresponding author
}
}

\usepackage[most]{tcolorbox}
\tcbuselibrary{breakable, skins}

\begin{document}
\maketitle
\begin{abstract}

Recent advances have enabled large language model (LLM) agents to solve complex tasks by orchestrating external tools. However, these agents often struggle in specialized, tool-intensive domains that demand long-horizon execution, tight coordination across modalities, and strict adherence to implicit tool constraints. Earth Observation (EO) tasks exemplify this challenge due to the multi-modal and multi-temporal data inputs, as well as the requirements of geo-knowledge constraints (spectrum library, spatial reasoning, etc): 
many high-level plans can be derailed by subtle execution errors that propagate through a pipeline and invalidate final results. A core difficulty is that existing agents lack a mechanism to learn fine-grained, tool-level expertise from interaction. Without such expertise, they cannot reliably configure tool parameters or recover from mid-execution failures, limiting their effectiveness in complex EO workflows. To address this, we introduce  \textbf{GeoEvolver}, a self-evolving multi-agent system~(MAS) that enables LLM agents to acquire EO expertise through structured interaction without any parameter updates. GeoEvolver decomposes each query into independent sub-goals via a retrieval-augmented multi-agent orchestrator, then explores diverse tool-parameter configurations at the sub-goal level. Successful patterns and root-cause attribution from failures are then distilled in an evolving memory bank that provides in-context demonstrations for future queries. Experiments on three tool-integrated EO benchmarks show that GeoEvolver consistently improves end-to-end task success, with an average gain of 12\% across multiple LLM backbones, demonstrating that EO expertise can emerge progressively from efficient, fine-grained interactions with the environment.

\end{abstract}

\section{Introduction}

\begin{figure}[t!]
  \centering
  \includegraphics[width=\columnwidth]{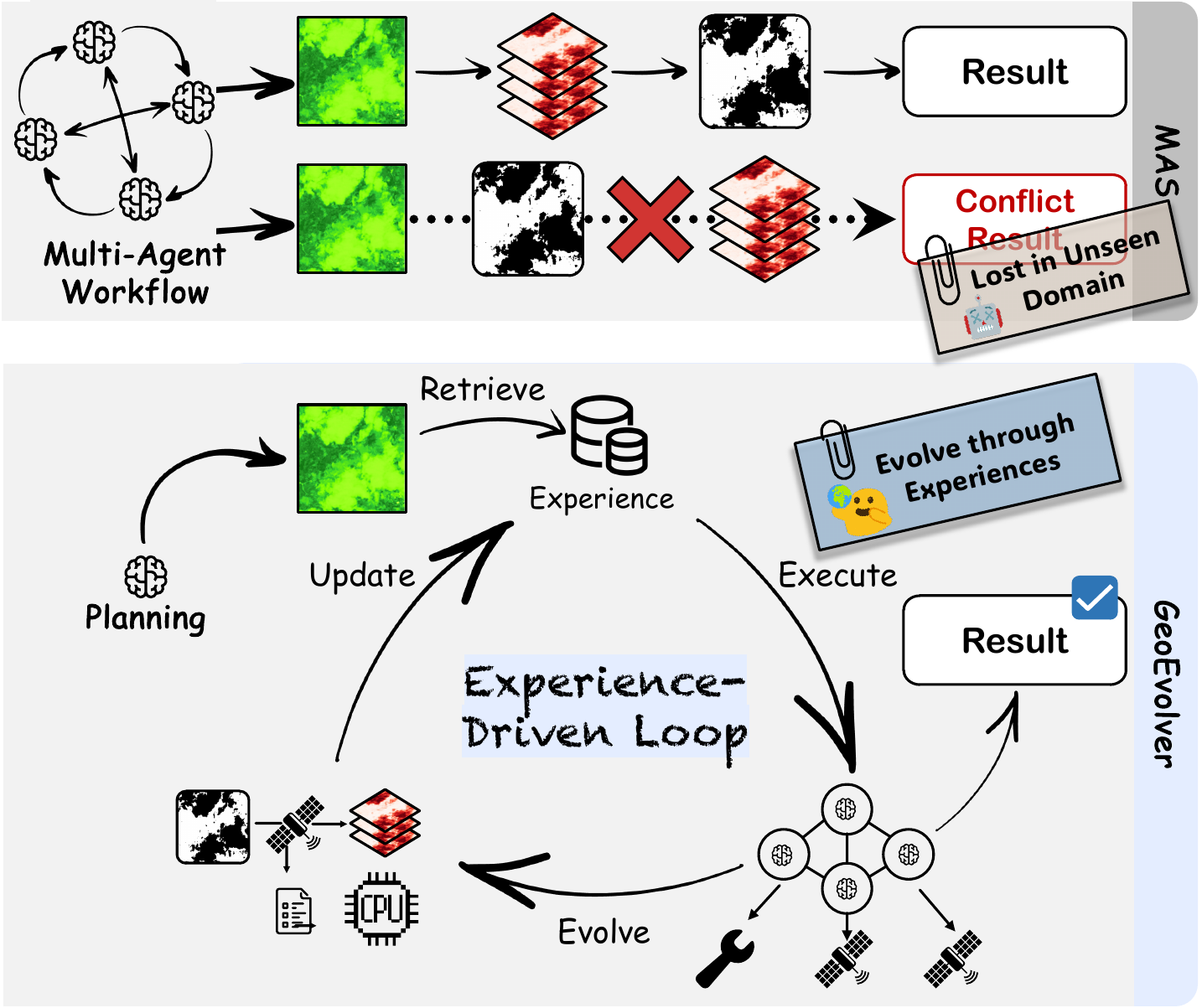}
  \caption{ Traditional MAS yield incomplete results on unseen tasks. \emph{GeoEvolver} utilizes an Experience-Driven Loop to continuously evolve memory, enabling parameter-free expertise acquisition.}
  \label{fig:single_col}
    \vspace{-3mm}
\end{figure}

LLM-based agents have demonstrated broad generality by orchestrating external tools and following language-based instructions across diverse domains.~\cite{wlflein2025llmagentsmakingagent, mialon2023gaiabenchmarkgeneralai,qin2023toolllm}.
However, when deployed in specialized domain workflows, these agents encounter a persistent bottleneck. Success depends not only on high-level planning but also on what we term \emph{execution groundedness}: the ability to align heterogeneous tool interfaces, track modality-specific states~(e.g., spatial resolutions or coordinate projections), and respect implicit physical constraints across long-term, multi-step trajectories~\cite{ding2025scitoolagent,chen2025iterresearch}. In such settings, small early mistakes can cascade into failures that are difficult to detect or recover from.


\par Earth Observation (EO) tasks ~\cite{xue2025regression} serve as a paradigmatic testbed for grounded scientific reasoning. Unlike pure text processing, EO workflows are state-coupled: the validity of a step depends not just on the input data, but on invisible metadata attributes like spatial resolution, temporal coverage, and coordinate projections. These attributes form a rigid physical context. Violating these couplings often leads to \emph{hallucinated execution}, where the toolchain runs successfully but the physical semantics are broken ~\cite{feng2025earth,xu2024rs}.

Despite recent progress, existing EO agents (as shown in Appendix~\ref{app:fail})~fall short of a general-purpose solution because they fail to bridge high-level reasoning with low-level precise execution. 
Language-centered models~\cite{shabbir2025thinkgeo,feng2025earth} can generate step-by-step plans, but they lack domain expertise to ground them in executable actions. Specifically, they struggle to configure precise tool parameters (e.g., projection thresholds), validate intermediate data states, or recover from execution failures~\cite{ding2025scitoolagent}. 
Vision-language models~\cite{kuckreja2024geochat,wang2025disasterm3,xuan2025dynamicvl}, by contrast, rely on extensive pretraining that is often decoupled from external toolchains. Ultimately, a structural gap remains: 
\textit{current agents operate with static knowledge, lacking a mechanism to acquire tool-level expertise from interaction or adapt to the vast, heterogeneous landscape of EO tools.}

\begin{table}[t]
\centering
\small
\renewcommand\arraystretch{0.80}
\setlength{\tabcolsep}{4.2pt}
\resizebox{\linewidth}{!}{%
\begin{tabular}{@{}ll c c l c@{}}
\toprule
\textbf{Domain} & \textbf{Method} & \textbf{\#Mod} & \textbf{\#Tools} & \textbf{Reasoning Type} & \textbf{Steps} \\
\midrule
\multirow{3.5}{*}{Chem} 
 & \begin{tabular}[t]{@{}l@{}}
     ChemCrow\\[-1pt]
     {\scriptsize\color{gray}\cite{bran2023chemcrow}}
   \end{tabular}
 & 2 & 18 & End-to-End & $\leq 11$ \\
 & \begin{tabular}[t]{@{}l@{}}
     ChemMCP\\[-1pt]
     {\scriptsize\color{gray}\cite{yu2025chemtoolagentimpacttoolslanguage}}
   \end{tabular}
 & 2 & 29 & End-to-End & $\leq 3$ \\
\midrule
\multirow{3.5}{*}{Bio}  
 & \begin{tabular}[t]{@{}l@{}}
     ProteinCrow\\[-1pt]
     {\scriptsize\color{gray}\cite{ponnapati2025proteincrow}}
   \end{tabular}
 & 3 & 36 & End-to-End & $\leq 15$ \\
 & \begin{tabular}[t]{@{}l@{}}
     ProtAgents\\[-1pt]
     {\scriptsize\color{gray}\cite{ghafarollahi2024protagentsproteindiscoverylarge}}
   \end{tabular}
 & 3 & 12 & End-to-End & $\leq 10$ \\
\midrule
\multirow{3.5}{*}{Med}  
 & \begin{tabular}[t]{@{}l@{}}
     MMedAgent\\[-1pt]
     {\scriptsize\color{gray}\cite{li2024mmedagentlearningusemedical}}
   \end{tabular}
 & 2 & 6  & End-to-End & $\leq 3$\\
 & \begin{tabular}[t]{@{}l@{}}
     MedAgent-Pro\\[-1pt]
     {\scriptsize\color{gray}\cite{wang2025medagent}}
   \end{tabular}
 & 3 & 4 & End-to-End & $\leq 12$ \\
\midrule
\multirow{7}{*}{\textbf{EO}}   
 & \begin{tabular}[t]{@{}l@{}}
     ThinkGeo\\[-1pt]
     {\scriptsize\color{gray}\cite{shabbir2025thinkgeo}}
   \end{tabular}
 & 3 & 14 & Step by Step &  $\leq 4$ \\
 & \begin{tabular}[t]{@{}l@{}}
     Earth-Agent\\[-1pt]
     {\scriptsize\color{gray}\cite{feng2025earth}}
   \end{tabular}
 & 4 & 104 & Step by Step & $\leq 19$ \\
 & \begin{tabular}[t]{@{}l@{}}
     Earth-Agent-MAS\\[-1pt]
     {\scriptsize\color{gray}\cite{li2025designingdomainspecificagentshierarchical}}
   \end{tabular}
 & 1 & 104 & Step by Step & $\leq 20$ \\
 & \begin{tabular}[t]{@{}l@{}}
     {\color{gray}\textit{Gap Analysis}}
   \end{tabular}
 & \textit{High} & \textit{High} & \textit{Policy-oriented} & \textit{Complex} \\
\bottomrule
\end{tabular}%
}
\caption{
Landscape of tool-using agents across scientific domains.
}
\label{tab:agent_landscape}
\vspace{-5mm}
\end{table}

\begin{figure*}[t]
    \centering
    \includegraphics[width=\textwidth]{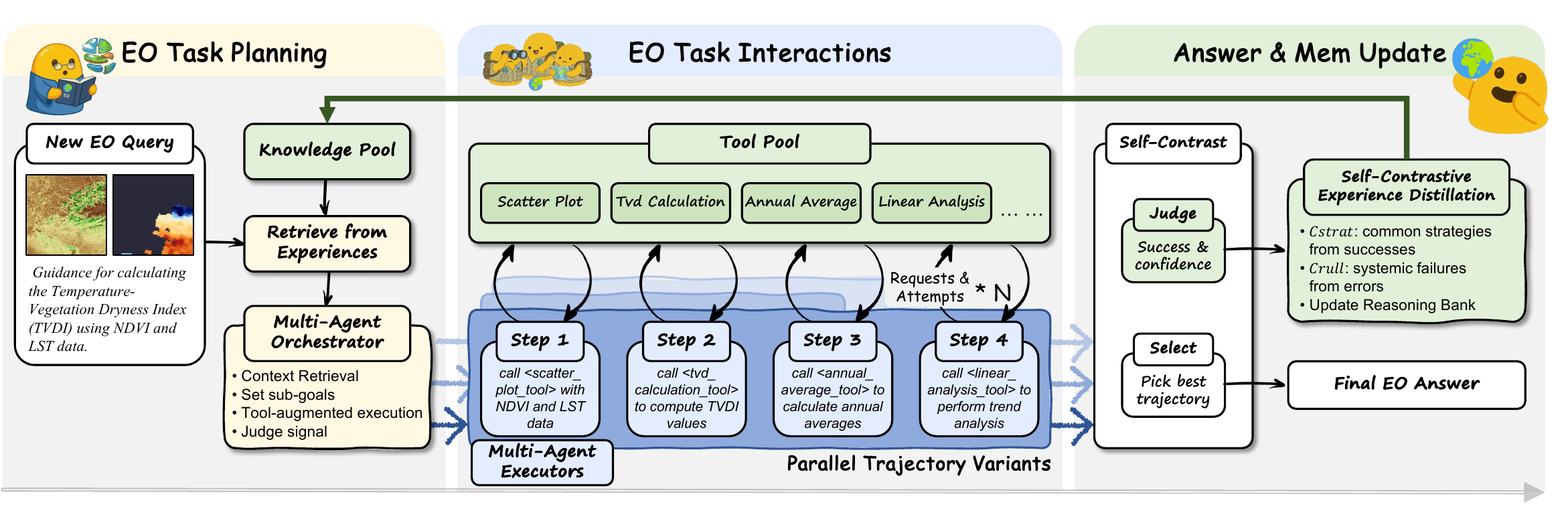}
    
    \caption{
        The overall architecture of GeoEvolver. 
        }
    \label{fig:main_arch}
\end{figure*}

\paragraph{Observation:} \emph{EO agent failures are dominated by execution-grounded errors rather than planning alone.} In EO workflows, geospatial primitives are physically grounded—coordinate reference systems, spatial resolution, spatial extent, and temporal coverage must remain consistent along the pipeline. LLM agents often treat these as symbolic parameters, which leads to plausible plans that nonetheless mis-handle projections, resampling, or temporal slicing. Such early, subtle mismatches silently propagate and derail downstream steps. Our analysis of prior EO agents (Appendix~\ref{app:fail}) shows that a large fraction of failures arises from execution-level mismatches and latent constraints, shifting the core challenge from ``how to plan'' to ``how to execute reliably under tool-coupled constraints ideally without parameter updates.''

\paragraph{Hypothesis}\emph{Can agents specialize without fine-tuning by accumulating verified interaction experience?}
Humans acquire domain competence not only through abstract planning, but also by accumulating execution-time priors~\cite{lake2017building} (e.g., tool failure modes, required constraints, and debugging strategies). Analogously, agents can interact with the environment at high frequency, verify constraint satisfaction, and distill reusable priors from both successful and failed trajectories \emph{without updating model parameters}. Recent advances in agent memory and retrieval-augmented reasoning indicate that storing distilled experience as retrievable in-context demonstrations can effectively shape future behavior~\cite{zhang2025agentic,ouyang2025reasoningbank,xu2025mem,zhao2024expelllmagentsexperiential,cai2025trainingfreegrouprelativepolicy}. However, existing memory mechanisms for general-purpose agents often emphasize planning-level knowledge and ignore the fine-grained, tool-level attribution and memorization crucial for EO, where such signals are costly to obtain in long-horizon tool use. This motivates our study of memory mechanisms tailored to EO.

To bridge planning and grounded tool execution, we propose \textbf{GeoEvolver}, a multi-agent framework that progressively acquires EO expertise through  fine-grained interaction \emph{without parameter updates}. GeoEvolver follows three principles: (1) \emph{Decomposition:} split each query into modular sub-goals with explicit I/O and dependencies to localize context and isolate errors; (2) \emph{Accumulation:} conduct query-adaptive yet stable high-frequency tool--environment interactions to collect diverse execution feedback; (3) \emph{Self-evolving:} distill these interactions into an evolving, retrievable memory bank that converts experience into domain priors. Our contributions are mainly three-fold:
\begin{itemize}[leftmargin=*,itemsep=0.2ex,topsep=0.2ex,parsep=0pt,partopsep=0pt]
\item \textbf{(Phenomenon)}
We identify a counter-intuitive finding compared with general agent design patterns: in EO domains, failures often arise not mainly from high-level plans, but from missing \textit{execution groundedness}---fine-grained tool expertise needed to satisfy implicit physical context constraints.
\item \textbf{(Mechanism)}
We propose \textbf{GeoEvolver}, an experience-driven multi-agent framework for EO tasks that decomposes queries, efficiently explores diverse tool configurations with environment feedback, and distills successes and failure patterns into an evolving memory bank as retrievable execution priors. 
\item \textbf{(Validation)}
On three tool-integrated EO benchmarks, GeoEvolver substantially improves end-to-end performance with an average gain of 12\% across multiple LLM backbones,  , showing that domain expertise can \textit{progressively emerge} from verification-driven environmental interactions rather than fine-tuning. 
\end{itemize}

\section{Related Work}
\paragraph{LLM Agents for Earth Observation (EO).}
Early LLM4EO work largely follows an end-to-end multimodal paradigm that aligns remote-sensing encoders with LLM embedding spaces for VQA~\cite{irvin2024teochat,liu2024remoteclip,kuckreja2024geochat}. 
Despite strong results on targeted tasks, this approach is costly to scale across EO’s heterogeneous sensors, spectral bands, and spatial resolutions, often requiring domain-specific encoders and repeated fine-tuning~\cite{wang2025disasterm3,xuan2025dynamicvl}. 
More recent efforts move toward tool-augmented agents that decouple perception from reasoning and execute EO workflows by orchestrating external tools (e.g., search APIs, raster processing, spatial analysis)~\cite{feng2025earth,shabbir2025thinkgeo,li2025designingdomainspecificagentshierarchical}. 
However, EO pipelines are highly tool-intensive and long-horizon: subtle parameter or format mistakes can propagate and invalidate downstream results. General-purpose tool-use ability learned from broad code/reasoning corpora is often misaligned with EO-specific tool constraints and usage patterns~\cite{ding2025scitoolagent,qin2023toolllm}, while supervised alignment requires expensive expert trajectories and scales poorly~\cite{xue2025regression}. 
GeoEvolver targets this gap by acquiring fine-grained, tool-level EO expertise from structured interaction, without parameter updates or manual trajectory annotation.

\paragraph{Multi-Agent Systems (MAS).}
MAS extend single-agent reasoning-and-acting (e.g., Chain-of-Thought, ReAct) by distributing roles and coordinating sub-tasks~\cite{wei2023chainofthoughtpromptingelicitsreasoning,yao2023reactsynergizingreasoningacting}. 
Many MAS designs rely on fixed workflows (e.g., debate, role-playing)~\cite{liang2024encouragingdivergentthinkinglarge,subramaniam2025multiagentfinetuningselfimprovement, li2025designingdomainspecificagentshierarchical} that are effective when the task type is stable, but become brittle as task diversity increases and new workflows require different coordination patterns. 
Automatic and dynamic orchestration methods improve flexibility by pruning or generating architectures~\cite{wang2025mas2selfgenerativeselfconfiguringselfrectifying,zhang2025aflowautomatingagenticworkflow}, yet their performance strongly depends on accurate task understanding and domain familiarity; when transferred to unseen, specialized domains outside pre-training distributions, agents often struggle to decompose problems and assign work effectively, especially with limited validation feedback. 
In contrast, GeoEvolver focuses less on searching new architectures and more on improving execution reliability via interaction-driven specialization at the sub-goal level.
\paragraph{Agent Memory and In-context Learning.}
To support long-horizon tasks, prior work explores memory through retrieval over interaction logs and hierarchical stores that separate working memory from long-term memory~\cite{zhang2025memgenweavinggenerativelatent,hua2025context}, as well as partitioning context across sub-agents~\cite{li2025intheflowagenticoptimizationeffective}. 
Beyond storage, several methods extract experience without fine-tuning, such as skill libraries, distilled insights, and reflection-based self-critiques~\cite{zhao2024expelllmagentsexperiential,ouyang2025reasoningbank,shinn2023reflexionlanguageagentsverbal}. 
A common limitation is granularity: experience is often summarized coarsely into lengthy narrative context, which is hard to reuse as a broadly applicable, tool-level strategy. Moreover, Existing approaches emphasize learning from success while under-utilizing failures~\cite{cai2025trainingfreegrouprelativepolicy}, losing diagnostic signals about \emph{why} execution breaks under implicit tool constraints. 
GeoEvolver tackles these challenges by leveraging extensive exploration and high-frequency, fine-grained interactions to learn transferable domain expertise, ensuring the memory context remains compact and efficient.

\section{Methods}

\subsection{Task Formulation}\label{sec:formulation}

We formulate EO workflow generation as a retrieval-conditioned sequential decision problem. Given a natural-language query $q$ and an initial environment state $e_0$ (e.g., available data sources and spatial context), the system generates a trajectory $\tau=(a_1,o_1,\dots,a_T,o_T)$, where $a_t$ denotes a tool action and $o_t$ the resulting observation.

\paragraph{Sub-goal Execution.}
Complex EO workflows are decomposed into $N$ sub-goals. Each sub-goal $g_n$ is assigned to a specialized subagent that independently executes a sub-trajectory $\tau_n$, transforming local inputs into the required intermediate artifacts. The segment-level outcome $Y_n \in \{0,1\}$ indicates whether subagent $n$ completes its sub-goal successfully. When execution fails, we record tool traces including error messages and logs, which are later used for failure attribution and experience accumulation.

\paragraph{Success Criterion.}
The overall task succeeds only if all sub-goals are achieved:
\begin{equation}
Y = \bigwedge_{n=1}^{N} Y_n.
\end{equation}
We denote the sub-goal success probability under subagent policy $\pi_n$ as $p_n(\cdot;\pi_n) = \Pr(Y_n{=}1 \mid g_n, \pi_n)$. The global objective is to find the optimal trajectory:
\begin{equation}
\tau^\star=\operatorname*{arg\,max}_{\tau}\Pr(Y{=}1\mid q,e_0;\mathcal{B}),
\end{equation}
where $\mathcal{B}$ is a non-parametric memory bank providing in-context guidance.

\paragraph{Design Goal.}
GeoEvolver aims to increase sub-goal success probabilities $p_n(\cdot)$ by reusing past interaction experience as retrievable priors, thereby improving the global success probability.

\subsection{Architecture}\label{sec:orch_exec}

As shown in Fig.~\ref{fig:main_arch}, GeoEvolver decouples high-level planning from low-level execution through role-based separation. The \emph{Orchestrator} decomposes queries into sub-goals and coordinates $N$ specialized \emph{Executors}, each responsible for certrain sub-tasks.

\paragraph{Role Specification.}
Each episode involves four operators that form a retrieve-plan-execute-judge closed-loop pipeline. Given an initial query $q$ and environment state $e_0$, the execution dynamics are:
\begin{equation}\label{eq:agent_dynamics}
\begin{aligned}
c &\leftarrow \pi_{\mathrm{retr}}(q;\mathcal{B})
&& \text{\footnotesize(Retrieval)} \\
\{g_n\}_{n=1}^{N} &\leftarrow \pi_{\mathrm{orch}}(q, c)
&& \text{\footnotesize(Orchestration)} \\
\tau_n &\leftarrow \pi_{\mathrm{exec}}^{(n)}(g_n) \quad \forall n \in [N]
&& \text{\footnotesize(Execution)} \\
Y, v &\leftarrow \mathcal{J}\bigl(\{\tau_n\}_{n=1}^{N}, q\bigr)
&& \text{\footnotesize(Judgement)}
\end{aligned}
\end{equation}

The \emph{Retriever} $\pi_{\mathrm{retr}}$ queries the memory bank $\mathcal{B}$ to construct a strategy context $c$ by aggregating relevant workflow templates and failure patterns (Sec.~\ref{sec:memory}). Conditioned on $(q, c)$, the \emph{Orchestrator} $\pi_{\mathrm{orch}}$ decomposes the query into $N$ sub-goals $\{g_n\}_{n=1}^{N}$, each specifying an interface contract (e.g., input/output formats and success criteria). The orchestrator then dispatches each sub-goal to a specialized \emph{Executor}. The $N$ subagents $\{\pi_{\mathrm{exec}}^{(n)}\}_{n=1}^{N}$ execute \emph{independently}: each receives only its assigned sub-goal $g_n$ and produces a sub-trajectory $\tau_n$ with no dependencies on other subagents. This independence enables parallel execution and localizes failure attribution. Finally, the \emph{Judge} $\mathcal{J}$ aggregates all sub-trajectories and emits a binary success label $Y$ together with auxiliary validity signals $v$ (e.g., format compliance, numeric matching). These signals inform both solution selection and memory updates (Sec.~\ref{sec:self_evolve}).

\paragraph{Parallel Exploration.}
EO workflows are tool-intensive and long-horizon, making single-pass execution error-prone. To improve robustness, GeoEvolver maintains $K$ parallel exploration variants $\{\mathcal{V}^{(k)}\}_{k=1}^{K}$. Each variant independently instantiates the pipeline above with prompt or decoding variations, producing a candidate solution $\tau^{(k)} = \{\tau_n^{(k)}\}_{n=1}^{N}$. Each variant may attempt up to $A$ corrective retries upon execution errors. The system selects the final solution as:
\begin{equation}
\hat{\tau}=\operatorname*{arg\,max}_{k\in[K]}\mathcal{S}(\tau^{(k)}\mid Y^{(k)},v^{(k)}),
\end{equation}
where $\mathcal{S}$ prioritizes verified success ($Y^{(k)}{=}1$) and uses validity signals $v^{(k)}$ as tie-breakers. The validity signal $v^{(k)}$ reflects a confidence estimate produced by $\mathcal{J}$, incorporating trajectory step count (penalizing verbose executions) and output correctness indicators.

\subsection{Hierarchical Memory}\label{sec:memory}

Long-horizon EO workflows require both persistent domain knowledge and episode-specific context. GeoEvolver addresses this with a two-tier memory architecture: a \emph{global} Memory Bank that accumulates reusable priors across episodes, and a \emph{local} Working Memory that tracks progress within each episode.

\paragraph{Memory Bank.} 
The Memory Bank $\mathcal{B}$ stores distilled procedural knowledge—reusable tool-chain patterns and failure-derived guardrails—that persists across queries. When a new query $q$ arrives, the retriever $\pi_{\mathrm{retr}}$ identifies relevant entries via embedding-based similarity. Let $f(\cdot)$ be an encoder mapping text to $L_2$-normalized vectors $\mathbf{e}$. The similarity between $q$ and a memory entry $m \in \mathcal{B}$ is the inner product $\operatorname{sim}(q, m) = f(q)^\top f(m)$. The retriever returns the top-$k$ entries $\mathcal{R}_k(q; \mathcal{B})$ and aggregates them into the strategy context:
\begin{equation}
c = \mathrm{Aggregate}\bigl(\mathcal{R}_k(q; \mathcal{B})\bigr).
\end{equation}
To prevent answer leakage during benchmark evaluation, retrieval applies a \emph{leakage filter} that excludes entries whose content co-occurs with expected outputs.

\paragraph{Working Memory.}
Within an episode, retaining the full interaction history quickly exceeds context limits. We maintain a compressed working context that balances global coherence with local detail:
\begin{equation}
H_t = \psi(H_{t-1}) \,\|\, \mathrm{tail}(\tau, L),
\end{equation}
where $\psi(\cdot)$ summarizes older interactions into a compact progress description, $\mathrm{tail}(\tau, L)$ retains the $L$ most recent raw steps, and $\|$ denotes concatenation.

\subsection{Experience-Driven Self-Evolution}\label{sec:self_evolve}

GeoEvolver improves over time by iteratively refining $\mathcal{B}$ from interaction experience. The evolution mechanism operates at two levels: (i) \emph{single-variant extraction} from the best solution, and (ii) \emph{contrastive distillation} across all exploration variants.

\paragraph{Single-Variant Extraction.}
For the selected best solution $\hat{\tau}$ (with outcome $\hat{Y}$), we extract structured knowledge via: 
\begin{equation}
m^\star = \mathcal{E}(\hat{\tau}, \hat{Y}),
\end{equation}
where $\mathcal{E}$ produces an \textbf{analysis pattern} if $\hat{Y}{=}1$ (capturing successful tool orderings and decision checkpoints) or an \textbf{error attribution} if $\hat{Y}{=}0$ (recording failure symptoms and corrective guardrails for future avoidance).

\paragraph{Contrastive Memory Distillation.}
To exploit the diversity from parallel exploration, we perform contrastive distillation across all $K$ variants. Rather than performing pairwise comparisons, we concatenate the candidate solutions $\{\tau^{(k)}\}_{k=1}^{K}$ along with their success labels into a unified contrastive block and prompt the LLM to synthesize transferable insights:
\begin{equation}
\mathcal{M}_{\mathrm{contrast}}=\mathrm{Distill}\bigl(q,\{(\tau^{(k)}, Y^{(k)})\}_{k=1}^{K}\bigr).
\end{equation}
The output $\mathcal{M}_{\mathrm{contrast}}$ contains memory items capturing workflow invariants across successful variants and recurring failure modes with corrective strategies.

\paragraph{Consolidation.}
Both single-variant and contrastive memories are integrated into the bank via key-based deduplication:
\begin{equation}
\mathcal{B} \leftarrow \mathcal{B} \cup \mathrm{Dedup}(\{m^\star\} \cup \mathcal{M}_{\mathrm{contrast}};\, \mathcal{B}),
\end{equation}
where $\mathrm{Dedup}(\cdot)$ filters out items whose canonical key $k(m) = (\texttt{source\_id}, \texttt{pattern\_type}, \texttt{title})$ already exists in $\mathcal{B}$. This deterministic deduplication keeps $\mathcal{B}$ compact while accumulating non-redundant procedural priors.

\section{Experiments}
\begin{table*}[!t]
    \centering
    \renewcommand\arraystretch{0.85}
    \setlength{\tabcolsep}{5pt}
    
    \resizebox{0.95\linewidth}{!}{%
    \begin{tabular}{@{}l@{\hspace{12pt}}
        c>{\columncolor{oursblue}}c@{\hspace{10pt}}
        c>{\columncolor{oursblue}}c@{\hspace{10pt}}
        c>{\columncolor{oursblue}}c@{\hspace{10pt}}
        c>{\columncolor{oursblue}}c@{\hspace{10pt}}
        c>{\columncolor{oursblue}}c@{}}
        
        \toprule

        \rowcolor{white}
        \multicolumn{11}{@{}c}{\dataset{Earth-Agent}} \\
        
        \rowcolor{white}
        \textbf{Method} &
        \multicolumn{2}{c}{\met{\textit{$Tool-A-O$}}{\uparrow}} &
        \multicolumn{2}{c}{\met{\textit{$Tool-I-O$}}{\uparrow}} &
        \multicolumn{2}{c}{\met{\textit{$Tool-E-M$}}{\uparrow}} &
        \multicolumn{2}{c}{\met{\textit{$Efficiency$}}{\downarrow}} &
        \multicolumn{2}{c}{\met{\textit{$Accuracy$}}{\uparrow}} \\
        \cmidrule(lr){2-3}\cmidrule(lr){4-5}\cmidrule(lr){6-7}\cmidrule(lr){8-9}\cmidrule(l){10-11}

        \rowcolor{white}
        & \small Base & \cellcolor{oursblueheader}\small\textbf{+Ours}
        & \small Base & \cellcolor{oursblueheader}\small\textbf{+Ours}
        & \small Base & \cellcolor{oursblueheader}\small\textbf{+Ours}
        & \small Base & \cellcolor{oursblueheader}\small\textbf{+Ours}
        & \small Base & \cellcolor{oursblueheader}\small\textbf{+Ours} \\
        \midrule

        \rowcolor{rowgray} GPT-5
         & 71.16 & 67.52\drop{}
         & 60.68 & 53.11\drop{}
         & 45.79 & 40.14\drop{}
         & 2.84  & \best{1.50}\imp{}
         & 63.16 & \best{70.85}\imp{} \\
        
        \rowcolor{white} Gemini-2.5
         & 61.80 & 61.43\drop{}
         & 50.78 & \best{53.60}\imp{}
         & 40.92 & 38.63\drop{}
         & 2.60  & \best{1.65}\imp{}
         & 55.06 & \best{59.91}\imp{} \\
        
        \rowcolor{white} GPT-4o
         & 66.88 & 62.71\drop{}
         & 53.20 & 48.93\drop{}
         & 47.47 & 37.45\drop{}
         & 2.61  & \best{1.56}\imp{}
         & 44.94 & \best{65.59}\imp{} \\
        
        \rowcolor{rowgray} DeepSeek-V3.1
         & 77.98 & 61.85\drop{}
         & 64.33 & 47.92\drop{}
         & 50.01 & 37.39\drop{}
         & 2.66  & \best{1.49}\imp{}
         & 52.23 & \best{59.68}\imp{} \\
        
        \rowcolor{white} Qwen3-32B
         & 42.39 & \best{62.84}\imp{}
         & 33.79 & \best{49.38}\imp{}
         & 26.10 & \best{42.32}\imp{}
         & 1.90  & \best{0.99}\imp{}
         & 24.80 & \best{46.96}\imp{} \\
        
        \rowcolor{avgrow} \textit{\small Avg.}
         & \multicolumn{2}{c}{}
         & \multicolumn{2}{c}{}
         & \multicolumn{2}{c}{}
         & \multicolumn{2}{c}{\impavg{+1.08}}
         & \multicolumn{2}{c}{\impavg{+12.56}} \\
  
        \arrayrulecolor{sectionline}\midrule\arrayrulecolor{black}

        \rowcolor{white}
        \multicolumn{11}{@{}c}{\dataset{ThinkGeo}} \\
        
        \rowcolor{white}
        \textbf{Method} &
        \multicolumn{2}{c}{\met{\textit{$Perception F1$}}{\uparrow}} &
        \multicolumn{2}{c}{\met{\textit{$Operation F1$}}{\uparrow}} &
        \multicolumn{2}{c}{\met{\textit{$Logic F1$}}{\uparrow}} &
        \multicolumn{2}{c}{\met{\textit{$Ans.$}}{\uparrow}} &
        \multicolumn{2}{c}{\met{\textit{$Ans~I$}}{\uparrow}} \\
        \midrule

        \rowcolor{rowgray}
        GPT-4o                   & \multicolumn{2}{c}{\best{87.05}} & \multicolumn{2}{c}{76.68} & \multicolumn{2}{c}{67.88} & \multicolumn{2}{c}{11.51} & \multicolumn{2}{c}{20.02} \\
        \rowcolor{white}
        GPT-4-1106               & \multicolumn{2}{c}{79.91} & \multicolumn{2}{c}{69.15} & \multicolumn{2}{c}{56.29} & \multicolumn{2}{c}{9.46} & \multicolumn{2}{c}{16.91} \\
        Claude-3.7-Sonnet        & \multicolumn{2}{c}{85.16} & \multicolumn{2}{c}{85.93} & \multicolumn{2}{c}{64.41} & \multicolumn{2}{c}{8.95} & \multicolumn{2}{c}{11.42} \\
        \rowcolor{rowgray}
        Qwen3-8B                 & \multicolumn{2}{c}{59.45} & \multicolumn{2}{c}{70.37} & \multicolumn{2}{c}{33.53} & \multicolumn{2}{c}{7.67} & \multicolumn{2}{c}{8.68} \\
        \rowcolor{white}

        \rowcolor{oursblue}
        \textbf{Ours (GPT-4o-mini based)}     & \multicolumn{2}{c}{83.90} & \multicolumn{2}{c}{\best{90.50}} & \multicolumn{2}{c}{\best{79.03}} & \multicolumn{2}{c}{\best{46.88}} & \multicolumn{2}{c}{\best{53.74}} \\

        \arrayrulecolor{sectionline}\midrule\arrayrulecolor{black}

        \rowcolor{white}
        \multicolumn{11}{@{}c}{\dataset{GeoPlan-Bench}} \\
        
        \rowcolor{white}
        \textbf{Method} &
        \multicolumn{2}{c}{\met{$Recall_{key}$}{\uparrow}} &
        \multicolumn{2}{c}{\met{$Precision_{key}$}{\uparrow}} &
        \multicolumn{2}{c}{\met{$F1_{key}$}{\uparrow}} &
        \multicolumn{2}{c}{\met{$Structural$}{\uparrow}} &
        \multicolumn{2}{c}{\met{$Holistic$}{\uparrow}} \\
        \midrule
        
        \rowcolor{white}
        ReAct~\cite{yao2023reactsynergizingreasoningacting}            & \multicolumn{2}{c}{0.33} & \multicolumn{2}{c}{0.50} & \multicolumn{2}{c}{0.37} & \multicolumn{2}{c}{0.47} & \multicolumn{2}{c}{962.57} \\
        \rowcolor{rowgray}
        AFlow~\cite{zhang2025aflowautomatingagenticworkflow}             & \multicolumn{2}{c}{0.39} & \multicolumn{2}{c}{0.62} & \multicolumn{2}{c}{0.44} & \multicolumn{2}{c}{\best{0.68}} & \multicolumn{2}{c}{992.60} \\
        \rowcolor{white}
        Earth-Agent-MAS~\cite{li2025designingdomainspecificagentshierarchical}    & \multicolumn{2}{c}{\best{0.66}} & \multicolumn{2}{c}{0.65} & \multicolumn{2}{c}{0.63} & \multicolumn{2}{c}{0.68} & \multicolumn{2}{c}{\best{1068.27}} \\
        
        \rowcolor{oursblue}
        \textbf{Ours}     & \multicolumn{2}{c}{0.64} & \multicolumn{2}{c}{\best{0.68}} & \multicolumn{2}{c}{\best{0.63}} & \multicolumn{2}{c}{0.45} & \multicolumn{2}{c}{1057.40} \\

        \bottomrule
    \end{tabular}%
    }
    \caption{Performance comparison across three benchmarks. \textit{Earth-Agent}: backbone ablation with our method. \textit{ThinkGeo \& GeoPlan-Bench}: comparison with existing methods.  \textcolor{impgreen}{$\blacktriangle$}\!: improvement.}
    \label{tab:main}
    \vspace{-4mm}
\end{table*}

In this section, we conduct a comprehensive evaluation to assess the effectiveness of GeoEvolver, aiming to address the following research questions (RQs):

\begin{itemize}[nosep, leftmargin=*]
    \item \textbf{RQ1:} How does GeoEvolver perform across diverse LLM backbones on the same EO analysis tasks?
    \item \textbf{RQ2:} How does GeoEvolver behave across LLM backbones of varying model capacities?
    \item \textbf{RQ3:} Can GeoEvolver maintain consistent performance across representative EO benchmarks with varying tool--modality coupling scenarios?
    \item \textbf{RQ4:} How does GeoEvolver perform compared with existing methods?
    \item \textbf{RQ5:} How sensitive is GeoEvolver to the number of executors ($N$), reasoning variants ($K$) as well as the retrieved memory items?
    \item \textbf{RQ6:} What is the contribution of each core component to the overall effectiveness of GeoEvolver?
\end{itemize}

\subsection{Experimental Setup}
\textbf{Dataset.}~~We evaluate on three EO agent benchmarks: \emph{ThinkGeo}~\cite{shabbir2025thinkgeo} (486 queries requiring multi-step spatial reasoning over optical/SAR imagery), \emph{EarthAgent}~\cite{feng2025earth} (236 tasks with structured workflows and tool dependencies), and \emph{GeoPlan-bench}~\cite{li2025designingdomainspecificagentshierarchical} (996 synthetic long-horizon planning tasks). 

\noindent
\textbf{Baselines.}~~We compare against three categories: (1) GeoEvolver instantiated with diverse scale LLM backbones (GPT, Gemini, DeepSeek, Qwen) to evaluate cross-backbone robustness; (2) state-of-the-art EO agents, multi-agent systems, and memory-based frameworks; 

\begin{table}[ht]
    \centering
    \renewcommand\arraystretch{1.00}
    \setlength{\tabcolsep}{4pt}
    
    \resizebox{\columnwidth}{!}{%
        \begin{tabular}{@{}lccccc@{}}
            \toprule
            \textbf{Method} & Tool-A-O & Tool-I-O  & Tool-E-M  & Eff.  & Acc. \\
            \midrule
            
            \rowcolor{white}
            \begin{tabular}[t]{@{}l@{}}
                Expel\\[-1pt]
                {\scriptsize\color{gray}\cite{zhao2024expelllmagentsexperiential}}
            \end{tabular}
                & 32.72\drop{} 
                & 25.94\drop{} 
                & 22.48\drop{} 
                & 1.79\drop{} 
                & 22.58\drop{} \\
            
            \rowcolor{rowgray}
            \begin{tabular}[t]{@{}l@{}}
                AFlow\\[-1pt]
                {\scriptsize\color{gray}\cite{zhang2025aflowautomatingagenticworkflow}}
            \end{tabular}
                & \best{58.07}\impavg{} 
                & 25.91\drop{} 
                & 22.38\drop{} 
                & \best{0.89}\impavg{} 
                & 30.16\drop{} \\
            
            \rowcolor{white}
            \begin{tabular}[t]{@{}l@{}}
                Deepagents\\[-1pt]
                {\scriptsize\color{gray}\cite{chase_langchain_2022}}
            \end{tabular}
                & 41.67\drop{} 
                & 33.98\drop{} 
                & 25.45\drop{} 
                & \second{1.06}\impavg{} 
                & 29.69\drop{} \\
            
            \rowcolor{rowgray}
            \begin{tabular}[t]{@{}l@{}}
                Training-free GRPO\\[-1pt]
                {\scriptsize\color{gray}\cite{cai2025trainingfreegrouprelativepolicy}}
            \end{tabular}
                & 57.24\drop{} 
                & \second{44.36}\drop{} 
                & \second{36.44}\drop{} 
                & 1.36\impavg{} 
                & \second{31.25}\drop{} \\
            
            \rowcolor{white}
            \begin{tabular}[t]{@{}l@{}}
                Earth-Agent-MAS\\[-1pt]
                {\scriptsize\color{gray}\cite{li2025designingdomainspecificagentshierarchical}}
            \end{tabular}
                & 32.28\drop{} 
                & 26.96\drop{} 
                & 20.91\drop{} 
                & 1.47
                & 15.87\drop{} \\
            
            \midrule
            \rowcolor{oursblue}
            \begin{tabular}[t]{@{}l@{}}
                \ours\\[-1pt]
            \end{tabular}
                & \second{57.66}
                & \best{44.66}
                & \best{39.06}
                & 1.47
                & \best{76.56} \\
            
            \bottomrule
        \end{tabular}%
    }
    \caption{Comparison with state-of-the-art methods on Earth-Agent Benchmark-subset-65.
    \best{Bold}: best. \second{Underline}: second best.}
    \label{tab:sota}
    \vspace{-5mm}
\end{table}

\paragraph{Implementation.}~Unless otherwise specified, we limit the number of executors $K$ to at most three, balancing execution diversity and computational cost in large-scale evaluation. 
The number of reasoning variants $N$ is capped at two to ensure stable comparison across settings. For experience retrieval from the Memory Bank, we consistently use top-$1$ retrieval throughout all experiments.

\paragraph{Evaluation Metrics.}~We follow the benchmarks setting to report end-to-end accuracy (whether the final output is correct) and step-by-step metrics (tool selection accuracy). Details are in the Appendix.

\begin{table}[!ht]
    \centering
    \renewcommand\arraystretch{1.00}
    \setlength{\tabcolsep}{4.5pt}
    \resizebox{\columnwidth}{!}{%
        \begin{tabular}{@{}lccccc@{}}
            \toprule
            \textbf{Variant} & \textbf{Tool-A-O} & \textbf{Tool-I-O} & \textbf{Tool-E-M} & \textbf{Eff.} & \textbf{Acc.} \\
            \midrule
            
            \rowcolor{oursblue}
            \begin{tabular}[t]{@{}l@{}}
                Full
            \end{tabular}
                & \best{57.66} & \best{44.66 } & \best{39.06 } & \best{1.47} & \best{76.56} \\
            
            \arrayrulecolor{sepline}\midrule\arrayrulecolor{black}
            
            \rowcolor{white}
            \begin{tabular}[t]{@{}l@{}}
                \wo{Self-contrast}\\[-1pt]
                {\scriptsize\color{gray}keep: MAS, Memory, Parallel}
            \end{tabular}
                & 61.07\impavg{} 
                & 48.30\impavg{} 
                & 36.60\drop{} 
                & 1.57\drop{} 
                & 54.69\drop{} \\
            
            \rowcolor{rowgray}
            \begin{tabular}[t]{@{}l@{}}
                \wo{Parallel Exploration}\\[-1pt]
                {\scriptsize\color{gray}keep: MAS, Memory}
            \end{tabular}
                & 41.67\drop{} 
                & 33.98\drop{} 
                & 25.45\drop{} 
                & 1.06\impavg{} 
                & 35.94\drop{} \\
            
            \rowcolor{white}
            \begin{tabular}[t]{@{}l@{}}
                \wo{Memory}\\[-1pt]
                {\scriptsize\color{gray}keep: MAS only}
            \end{tabular}
                & 57.68\impavg{} 
                & 27.74\drop{} 
                & 24.50\drop{} 
                & 0.99\impavg{} 
                & 31.25\drop{} \\
            
            \arrayrulecolor{sepline}\midrule\arrayrulecolor{black}
            
            \rowcolor{rowgray}
            \begin{tabular}[t]{@{}l@{}}
                Baseline\\[-1pt]
                {\scriptsize\color{gray}no MAS / Memory / Parallel / Failure Replay}
            \end{tabular}
                & 70.26\impavg{} 
                & 48.05\impavg{} 
                & 39.97\impavg{} 
                & 1.79\drop{} 
                & 25.00\drop{} \\
            
            \bottomrule
        \end{tabular}%
    }
    
    \caption{Ablation study on Earth-Agent Benchmark-subset-65. \impavg: improvement from full model. \drop: decrease from full model.}
    \label{tab:ablation}
    \vspace{-2mm}
\end{table}

\subsection{Analysis}
\subsubsection{Quantitative Study}

\begin{figure}[t]
    \centering
    \includegraphics[width=0.75\linewidth]{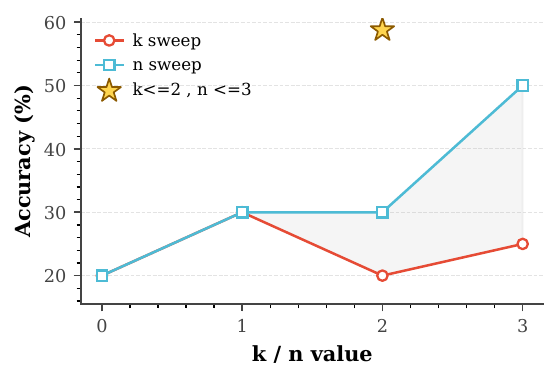}
    \caption{Sensitivity analysis of hyper-parameters $k$ and $n$}
    \label{fig:sensitivity}
    \vspace{-5mm}
\end{figure}

\begin{figure}[t]
    \centering
    \includegraphics[width=0.75\linewidth]{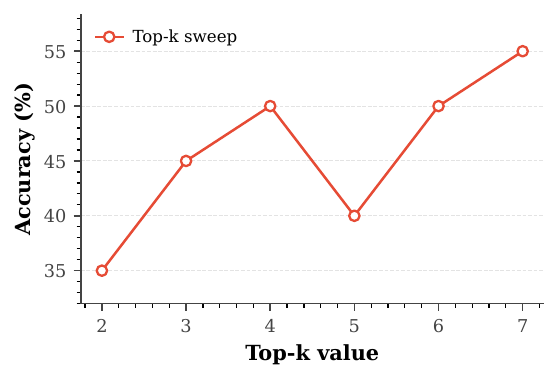}
    \caption{Sensitivity analysis of the number of retrieved memory items}
    \label{fig:sensitivity_topk}
    \vspace{-3mm}
\end{figure}

\paragraph{Cross-Backbone Generalization (RQ1).}
We first assess whether GeoEvolver generalizes across LLM backbones.
As shown in Table~\ref{tab:main} (Earth-Agent), GeoEvolver delivers consistent gains over all five foundation models, improving end-to-end accuracy by an average of $+12.56$ percentage points (pp).
Notably, the effect is strongest for smaller models: Qwen3-32B increases from $24.80\%$ to $46.96\%$ ($+22.16$pp), suggesting that experience accumulation can partially compensate for limited parametric capacity.
These results confirm that domain expertise can emerge through structured interaction and experience reuse, independent of model scale.

\paragraph{Step-Level vs.\ Task-Level Metrics.} A striking pattern emerges: step-wise metrics occasionally \emph{decrease} after applying GeoEvolver (e.g., GPT-5 Tool-A-O: $71.16\%\rightarrow67.52\%$), yet end-to-end accuracy improves substantially ($63.16\%\rightarrow70.85\%$). This motivates our core design insight: \textit{LLM-generated tool chains can deviate from human-annotated trajectories while remaining functionally correct}. Rather than imitating demonstrations---which risks brittle adherence to specific trajectories---GeoEvolver discovers \emph{functionally equivalent} paths that satisfy task constraints. The consistent decoupling between step-level and task-level metrics across main results and ablations confirms that trajectory imitation is unsuitable for domain adaptation in tool-rich EO settings.
\paragraph{Scaling with Model Capacity (RQ2).}
To isolate the effect of model capacity, we compare open-source models spanning different scales.
Qwen3-32B shows the largest relative improvement ($+89.4\%$), whereas the stronger DeepSeek-V3.1 still gains meaningfully ($+14.3\%$).
This pattern indicates that experience-driven adaptation benefits models across the capacity spectrum, while smaller models benefit disproportionately from externalized experience and memory that augment their limited capacity.
\paragraph{Cross-Benchmark Consistency (RQ3).} We evaluate GeoEvolver on three benchmarks. On \textit{ThinkGeo}, which requires multi-step spatial reasoning, GeoEvolver (GPT-4o-mini) reaches $46.88\%$ accuracy---$4.1\times$ over GPT-4o ($11.51\%$)---with $90.50\%$ Operation-F1 and $79.03\%$ Logic-F1. On \textit{GeoPlan-Bench} (996 synthetic planning tasks), GeoEvolver matches Earth-Agent-MAS on $\mathrm{F1}_{\text{key}}$ while improving $\mathrm{Precision}_{\text{key}}$ ($0.68$ vs.\ $0.65$); the narrower gap reflects that execution complexity is intentionally low. On \textit{Earth-Agent}, the most challenging benchmark with 104 heterogeneous tools, GeoEvolver achieves the largest gains ($+12.56$pp), confirming that experience-driven adaptation is most valuable when grounded execution is the primary bottleneck.

\subsubsection{Comparison with Existing Methods (RQ4)}

Table~\ref{tab:sota} compares GeoEvolver with representative memory frameworks and multi-agent systems on Earth-Agent Benchmark.
\paragraph{Memory-Based Methods.}
Expel and Training-free GRPO underperform substantially ($22.58\%$ and $31.25\%$ vs.\ $76.56\%$).
We attribute this to two core limitations:
(1) \textit{coarse-grained execution}---operating primarily at the trajectory level makes failures hard to localize and correct;
(2) \textit{loosely structured memory}---retrieved experiences are not compressed into actionable, decision-ready units and may introduce irrelevant context that distracts execution.
These results suggest that effective memory for tool use requires both fine-grained organization and structured representations that directly support planning and execution.
\paragraph{Multi-Agent Systems.}
Earth-Agent-MAS relies on a \textit{fixed workflow} tuned for specific templates and thus degrades sharply under diverse queries, achieving only $15.87\%$ accuracy.
AFlow introduces \textit{flexible architecture search} and improves generality ($30.16\%$) but still struggles with out-of-distribution EO tasks.
These findings indicate that architectural flexibility alone is insufficient to internalize domain-specific tool constraints and interface conventions.
By combining \textit{adaptive topology} with \textit{domain-grounded memory}, GeoEvolver attains $76.56\%$ accuracy, leveraging accumulated EO expertise while dynamically adjusting its structure to the task on-the-fly.

\subsubsection{Ablation Study}

\paragraph{Component Contribution (RQ6).}
Table~\ref{tab:ablation} reports progressive ablations starting from the full model ($76.56\%$).
(1) Removing Self-contrast reduces accuracy to $54.69\%$ ($-21.87$pp), highlighting the importance of failure-driven diagnostic signals.
(2) Further removing Parallel Exploration drops performance to $35.94\%$ ($-18.75$pp), indicating that diverse trajectory sampling substantially increases the probability of finding successful and informative executions.
(3) Removing Memory yields $31.25\%$ ($-4.69$pp), confirming that persistent experience storage provides additional gains beyond exploration alone.
(4) The single-agent baseline attains only $25.00\%$.
Overall, GeoEvolver provides a cumulative improvement of $+51.56$pp over the single-agent baseline, with Self-contrast and Parallel Exploration contributing the largest individual gains.
Importantly, the ablation variants exhibit the same step-level vs.\ task-level decoupling observed in the main experiments, further supporting our design rationale.

\paragraph{Hyperparameter Sensitivity (RQ5).}
Figure~\ref{fig:sensitivity} examines sensitivity to the number of parallel variants $K$ and  executors $N$.
We observe a clear \textit{interaction effect}: tuning either in isolation yields limited gains, whereas jointly increasing both produces substantial improvements.
This arises from complementary roles: decomposition granularity ($N$) determines whether sub-problems are fine-grained enough to expose tool-specific constraints, while exploration breadth ($K$) determines whether sufficient trajectory diversity exists to surface informative feedback.
When $N$ is low, increasing $K$ leads to diminishing returns—parallel variants explore similar action spaces and produce redundant outcomes.
When $N$ is high, increasing $K$ amplifies gains, as diverse exploration over fine-grained sub-goals yields precise, reusable priors.
In practice, moderate settings ($K \leq 2$, $N \leq 3$) strike an effective balance between performance and computational cost.
Figre~\ref{fig:sensitivity_topk} examines sensitivity to the number of memory items top-k. Performance scales positively with memory size, demonstrating that even a compact memory structure can supply the model with effective contextual guidance.

\section{Discussion}
\label{sec:discussion}
\paragraph{Parametric vs.\ Non-Parametric Domain Specialization.}
Our results challenge the assumption that effective domain specialization requires computationally expensive fine-tuning. We show that a general-purpose LLM, augmented with structured experience accumulation and retrieval, can match or surpass domain-specialized models. This suggests an alternative design pattern we term \emph{Experience-as-Parameters}: rather than encoding domain knowledge into static model weights vulnerable to catastrophic forgetting, we externalize it into a dynamic \textit{Memory Bank}. This approach is particularly suited for EO, where sensor characteristics and analytical tools evolve rapidly—updating interaction traces is substantially cheaper than retraining foundation models.
\paragraph{Experience from Exploration as Contextual Evidence.}
Our ablations reveal that actionable signals emerge not only from successful rollouts but from the exploration process itself. In scientific tool-use workflows, each interaction—whether successful or failed—produces contextual evidence about the environment's feasible region. By distilling both validated procedures and failure-induced constraints into memory, GeoEvolver rules out invalid actions early, shrinking the search space and preventing recurring mistakes. This mirrors expert practice, where understanding boundary conditions often provides decisive guidance when correct procedures are underspecified.

\section*{Limitations}
Despite the observed gains, our approach has several limitations. First, inference latency and cost remain a concern: the Parallel Exploration mechanism (with $K$ variants) increases token usage and wall-clock time approximately linearly in $K$. While this is acceptable for offline analysis, it can be prohibitive in time-critical settings (e.g., rapid disaster response), and the accuracy--latency trade-off as a function of $K$ merits more systematic modeling. Second, the system exhibits a ``Blind Agent'' limitation. Unlike native vision-language models, our tool-use agents operate on images primarily through metadata and file handles, and do not directly process pixel data during intermediate reasoning. This abstraction can reduce sensitivity to high-dimensional visual noise, but it also limits the agent's ability to perform tasks that benefit from direct visual inspection (e.g., diagnosing subtle artifacts in intermediate rasters). Third, the framework depends on the reliability of the external toolchain. GeoEvolver acts as an orchestrator rather than a numerical engine; consequently, its performance is bounded by the correctness and robustness of the underlying EO tools (e.g., GDAL). If a tool fails silently or outputs physically implausible results without explicit errors, the current agent has limited grounding to detect such semantic anomalies.
\paragraph{Future Directions.}
We identify three extensions that merit exploration: (1) \textit{Experience distillation}—fine-tuning compact SLMs on Memory Bank traces to compress ``System~2'' reasoning into efficient ``System~1'' policies; (2) \textit{Hybrid vision--tool architectures}—augmenting the Judge with lightweight visual encoders for intermediate sanity checks; (3) \textit{On-board deployment}—adapting the framework for edge satellites under bandwidth and compute constraints.


\bibliography{custom}

\appendix
\onecolumn

\section{Appendix}
\label{sec:appendix}

\subsection{Generative AI statement}
This work employed Open AI GPT-5 for language-related assistance, including manuscript polishing, clarity improvement, and grammatical refinement. All scientific content, technical claims, experimental results, and interpretations were authored by the researchers and have been thoroughly reviewed and validated by humans. The use of GPT-5 did not involve the generation of new scientific ideas, experimental designs, data analyses, or conclusions, and did not replace expert judgment at any stage of the research process.

\paragraph{Potential Risks}
A potential risk in this study arises from the use of external remote sensing analysis tools and AI-assisted systems, which may exhibit hallucinations or spurious inferences, particularly when operating under domain shifts, incomplete metadata, or ambiguous geospatial contexts. Such tools may produce outputs that appear plausible but are not physically or scientifically grounded. Nevertheless, residual uncertainties related to tool-induced hallucinations cannot be entirely eliminated and should be considered when deploy this method to real world.

\subsection{Algorithm}
\begin{algorithm}[ht]
\caption{Pipeline}
\label{alg:GeoEvolver}
\small
\begin{algorithmic}[1]

\Require Query set $\mathcal{Q}$; tool suite $\mathcal{T}$; long-term experience memory $\mathcal{M}_{\mathrm{LTM}}$; parallel budget $N$; max executor $K_{\max}$
\Ensure  Per-query best trajectories $\mathcal{R}^*$; updated experience memory $\mathcal{M}_{\mathrm{LTM}}$
\State \textit{Objective:} maximize task success and improve the quality of distilled interaction experience stored in $\mathcal{M}_{\mathrm{LTM}}$ over $\mathcal{Q}$.

\Procedure{GeoEvolver}{$\mathcal{Q}, \mathcal{T}, \mathcal{M}_{\mathrm{LTM}}, N, K_{\max}$}
  \State $\mathcal{R}^* \gets \emptyset$
  \For{$q \in \mathcal{Q}$}

    \Statex \Comment{\textbf{Phase 1: Experience-Guided Multi-Agent Planning}}
    \State $\mathcal{E}_q \gets \Call{Retrieve}{\mathcal{M}_{\mathrm{LTM}}, q}$ \Comment{retrieve relevant experiences}
    \State $\pi \gets \Call{Plan}{q, \mathcal{E}_q}$ \Comment{construct segmented multi-agent plan}
    \State $\kappa \gets \min(\Call{Scale}{\pi}, K_{\max})$ \Comment{determine \#segments}
    \State $\mathcal{W} \gets \Call{InitWM}{\kappa}$ \Comment{Initialize work state for the $\kappa$ subagents}
    \State $\mathbb{T}_q^+ \gets \emptyset,\;\; \mathbb{T}_q^- \gets \emptyset$ \Comment{successful / failed trajectory sets}

    \Statex \Comment{\textbf{Phase 2: Experience-Conditioned Parallel Execution }}
    \For{$i \gets 1$ \textbf{to} $N$ \textbf{in parallel}}
      \State $q_i \gets \Call{Diversify}{q, i}$ \Comment{create i.i.d.\ diversified query variant / seed}
      \State $\tau_i \gets \Call{Agent}{q_i, \mathcal{T}, \mathcal{W}, \mathcal{E}_q}$ \Comment{multi-agent tool trajectory across $\kappa$ segments}
      \State $\hat{y}_i \gets \Call{Parse}{\tau_i}$ \Comment{parse final answer from trajectory}
      \State $s_i \gets \mathbb{I}[\hat{y}_i = q.\texttt{judge}]$ 
      \State $\gamma_i \gets \Call{Confidence}{\tau_i}$ \Comment{trajectory-level confidence score}
      \State $r_i \gets \alpha s_i + \beta \gamma_i - \lambda |\tau_i|$ \Comment{reward: success \& confidence vs.\ length penalty}
      \If{$s_i = 1$}
        \State $\mathbb{T}_q^+ \gets \mathbb{T}_q^+ \cup \{\tau_i\}$
      \Else
        \State $\mathbb{T}_q^- \gets \mathbb{T}_q^- \cup \{\tau_i\}$
      \EndIf
    \EndFor

    \Statex \Comment{\textbf{Phase 3: Contrastive Experience Update}}
    \State $i^* \gets \arg\max_{i \in \{1,\dots,N\}} r_i$ \Comment{parallel selection of best trajectory}
    \State $\tau^* \gets \tau_{i^*}$
    \State $\mathcal{E}_q^{\text{new}} \gets \Call{ExtractContrast}{\mathbb{T}_q^+, \mathbb{T}_q^-, q}$ \Comment{build contrastive pairs $(\tau^+,\tau^-)$}
    \State $\mathcal{M}_{\mathrm{LTM}} \gets \mathcal{M}_{\mathrm{LTM}} \cup \mathcal{E}_q^{\text{new}}$ \Comment{append distilled experience}
    \State $\mathcal{R}^* \gets \mathcal{R}^* \cup \{(q, \tau^*)\}$
    \State \Call{Discard}{$\mathcal{W}$} \Comment{drop task-local work state; long-term memory only in $\mathcal{M}_{\mathrm{LTM}}$}

  \EndFor
  \State \Return $\mathcal{R}^*, \mathcal{M}_{\mathrm{LTM}}$
\EndProcedure
\end{algorithmic}
\end{algorithm}

\newpage
\twocolumn
\subsection{Main Evaluation Metrics Detail}

\textbf{1. Accuracy $\uparrow$}
\begin{equation}
\text{2. Acc} = \mathbb{E}_{g \sim \mathcal{G}}\left[\mathbb{I}\{y = y^*\}\right]
\end{equation}

\noindent
\textbf{3. Efficiency $\downarrow$}
\begin{equation}
\text{Eff}(\tau, \tau^*) = \frac{|\tau|}{|\tau^*|}
\end{equation}

\noindent
\textbf{4. Tool-Any-Order (T-A-O) $\uparrow$}
\begin{equation}
\text{TAO}(\tau, \tau^*) = \frac{|\text{Set}(t^*) \cap \text{Set}(t)|}{|\text{Set}(t^*)|}
\end{equation}

\noindent
\textbf{5. Tool-In-Order (T-I-O) $\uparrow$}
\begin{equation}
\begin{split}
k^* = \max\big\{k : \exists\, 1 \leq j_1 < \cdots < j_k \leq n,\\
t_{j_i} = t_i^*,\; \forall i \leq k\big\}
\end{split}
\end{equation}

\begin{equation}
\text{TIO}(\tau, \tau^*) = \frac{k^*}{m}
\end{equation}

\textbf{6. Tool-Exact-Match (T-E-M) $\uparrow$}
\begin{equation}
\ell_{\text{lcp}} = \max\left\{\ell \leq \min(m, n) : t_i = t_i^*,\; \forall i \leq \ell\right\}
\end{equation}
\begin{equation}
\text{TEM}(\tau, \tau^*) = \frac{\ell_{\text{lcp}}}{m}
\end{equation}

\textbf{Notation}

\begin{enumerate}[leftmargin=1em, labelsep=0.5em, itemsep=0pt]
    \item $\tau^*$: Expert-annotated ground-truth trajectory
    \item $\tau$: Agent-predicted trajectory
    \item $m$: Length of ground-truth trajectory
    \item $n$: Length of predicted trajectory
    \item $t^* = (t_1^*, \ldots, t_m^*)$: Ground-truth tool sequence
    \item $t = (t_1, \ldots, t_n)$: Predicted tool sequence
    \item $y^*$: Ground-truth final answer
    \item $y$: Predicted final answer
    \item $\mathcal{G}$: Distribution of benchmark tasks
    \item $\mathbb{I}\{\cdot\}$: Indicator function
    \item $\text{Set}(\cdot)$: Function extracting the set of unique tools
\end{enumerate}

\newpage
\onecolumn

\subsection{Prompt Template}
\label{sec:appendix_prompts}
\begin{promptbox}{orchestrator system prompt}
You orchestrate a team of remote-sensing subagents (retriever, executors, judge) to answer multiple-choice questions about Earth observation data.

Your responsibilities:

1. Pull relevant strategy memories, plan at a high level, and delegate concrete tool work to the executors.

2. Monitor their progress, ask for clarifications or adjustments when workflow guardrails are violated, and prefer delegation over direct low-level tool use.

3. Keep the discussion coherent by narrating plan, tool intent, and observed outcomes in plain text so the log captures reasoning.

4. At the end of every problem produce:
   <Diag>
   
   Tool summary: <bullet list describing each tool call and its intent/result>.
   
   Failure reason: <if something failed, explain root cause; otherwise 'None'>.
   
   </Diag>
   
5. Finish with the final answer line `<Answer>Your choice</Answer>` using the letter of the selected option.

6. Enforce any directory hints or path constraints from the user prompt by reminding executors to obey them.
"""
\end{promptbox}

\begin{promptbox}{variants act  prompt}

You are solving a geoscience/remote-sensing task. Use step-by-step reasoning, tool-free.
Incorporate strategies below strictly as high-level hints (do NOT copy numeric thresholds, coordinates, or pixel values).
Focus on: tool selection, parameter tuning, data preprocessing (cloud/shadow masks, reproject/resample), QA filtering, temporal alignment, and validation.

{memory\_block}
{diversity\_hint}Task: {query}
Answer with both reasoning and a final answer line like 'ANSWER: ...' when applicable. If producing metrics, explain how they were derived conceptually rather than copying numbers.
"""

\end{promptbox}

\begin{promptbox}{executor}
    "You are a geoinformatics executor who follows strategist instructions exactly."
    
    "- Use only file paths provided by the prompt or previous tool outputs (benchmark/data/... or tool results)."
    
    "- Reuse absolute output paths returned by tools; never invent directories."
    
    "- If a directory contains many files, list them in pages using `ls(path=..., offset=0, limit=50)` and advance offset in later calls."
    
    "- Keep a action result narrative. Work memory is updated automatically; only call update work memory manually if you need to adjust the plan significantly."
    
    "- Prefer batch tools for repeating rasters, validate paths with ls/glob, and pass JSON arguments explicitly."
    
    "- If a tool fails, summarize the error, adjust inputs once, and retry only when justified."
    
    "- Respect temporal/spatial filters, base proportions on actual pixel counts, and avoid emitting <Answer> until every checklist item is resolved."

"""
\end{promptbox}

\begin{promptbox}{judge}
You are a strict evaluator for remote-sensing reasoning trajectories. You do NOT know the gold answer. Judge only internal consistency, tool usage correctness, parameter validity, and whether the final answer is supported by the trajectory. If the response lacks an explicit final answer tag, mark FAILURE.

"tools": [judge\_tool, extract\_tool]

Question:
{query\_text}

Final answer:
{final\_text}

Reasoning:
{reasoning\_text}

Tool trace:
{tool\_trace}

Diagnostics:
{diag\_text or 'N/A'}

Respond with JSON: {"decision": "SUCCESS" | "FAILURE", "confidence": 0-1, "justification": "..."}.

"""

\end{promptbox}

\begin{promptbox}{extractor}
You are analyzing multiple geoscience/remote-sensing trajectories to extract reusable ANALYSIS PATTERNS and ERROR ATTRIBUTIONS.

Return ONLY valid JSON following this schema (no prose outside JSON):

\{

  "memories": [
  
    \{
    
      "title": "...",
      
      "description": "...",
      
      "content": "Remote-sensing guidance that spells out the ordered tool-chain (Tool A -> Tool B -> Tool C) and key decision checkpoints",
      
      "pattern\_type": "analysis\_pattern" OR "error\_attribution
      ",
      
      "action\_items": ["Step 1: call <tool\_name> ...", "Step 2: ..."],
      
      "detection\_cues": ["Signals to apply/watch for this workflow"],
      
      "failure\_cause": "Root cause + tool error hint (only for error\_attribution)"
      
    \}
    
  ]
  
\}

Extract 2-4 total entries across successes and failures. Avoid numeric leakage; emphasize sequential tool usage, QA/cloud masking, reprojection/resampling, temporal alignment, unit conversions, validation, and quote representative error messages when applicable.

QUERY: \{query\}

\{trajectories\_block\}

"""

\end{promptbox}

\subsection{Case Study}
\subsubsection{Result cases}
\begin{promptbox}{case1: Long-chain Success Case}
\textbf{Question ID:} 34 \\
\textbf{Task:} Split-window LST (Band31/Band32) $\rightarrow$ compute annual mean LST for 2018--2023 (Guangzhou) $\rightarrow$ select the correct option.

\textbf{Data:} Band31/Band32 brightness temperature TIFFs + emissivity files (Emis31, Emis32)

\textbf{Why this is a good showcase:} Very long multi-step execution (149 LST generations + 6 yearly aggregations) while staying consistent and correct.

\textbf{Plan highlights:}
\begin{enumerate}[nosep,leftmargin=*]
    \item List files in data directory to find Band31/Band32 TIFFs
    \item Pair Band31 and Band32 files by acquisition date and plan LST output paths
    \item Run \texttt{split\_window} on each pair to produce daily LSTs
\end{enumerate}

\textbf{Execution Statistics:}
\begin{itemize}[nosep,leftmargin=*]
    \item \texttt{tool\_calls} = 166, shown = 155 (non-OS), \texttt{unique\_tools} = 2
    \item 333 trace entries (call/result)
    \item Auxiliary tools: \texttt{write\_todos} $\times$2, \texttt{ls} $\times$2, \texttt{glob} $\times$7
    \item Core tools: \texttt{split\_window} $\times$149, \texttt{calc\_batch\_image\_mean\_mean} $\times$6
\end{itemize}

\textbf{Input Pairing:} Each \texttt{split\_window} call used (BT\_31, BT\_32) + emissivity (Emis31, Emis32) matched by timestamp (e.g., \texttt{2019\_01\_01\_0250}), writing LST to per-date output paths.

\textbf{Aggregation:} \texttt{glob} collected yearly LST subsets by filename year prefix; \texttt{calc\_batch\_image\_mean\_mean} computed per-year mean-of-means. Max--min spread $\approx$ 12.46\,K (2019 vs.\ 2023).

\textbf{Key Intermediate Results (Annual Mean LST, K):}
\begin{center}
\begin{tabular}{cccccc}
\toprule
2018 & 2019 & 2020 & 2021 & 2022 & 2023 \\
\midrule
294.36 & 297.86 (max) & 296.10 & 288.89 & 287.63 & 285.41 \\
\bottomrule
\end{tabular}
\end{center}

\textbf{Final Answer:} \texttt{<Answer>B</Answer>} \quad (GT = B, \textcolor{green!60!black}{\textbf{Correct}})
\end{promptbox}

\begin{promptbox}{case2: Recovery with Memory Bank}
\textbf{Question ID:} 27 \\
\textbf{Task:} ASTER TIR (Bands 10--12) TTM $\rightarrow$ compute $\Delta$LST between two regions (Dec 23, 2022) $\rightarrow$ select the correct option.

\textbf{Data:} ASTER TIR Bands 10--12 brightness temperature files for two polygons

\textbf{Baseline Error:} \texttt{Error calling tool `temperature\_emissivity\_separation': list index out of range}

\textbf{Strategy Hint:} ASTER TIR LST TTM Processing Chain

\textbf{Plan highlights:}
\begin{enumerate}[nosep,leftmargin=*]
    \item List files in data directory to find ASTER TIR Bands 10--12 for 2022-12-23 (10:30 AM)
    \item Group files into TIR band triplet (Bands 10, 11, 12) for the acquisition and verify paths
    \item Run \texttt{temperature\_emissivity\_separation} on verified TIR band paths (use \texttt{representative\_band\_index=2}) to produce LST
\end{enumerate}

\textbf{Recovery Detail:} The initial failure came from invalid band list/indexing; the retry explicitly verified BT\_10/11/12 paths per polygon and passed \texttt{representative\_band\_index=2}.

\textbf{Execution Statistics:}
\begin{itemize}[nosep,leftmargin=*]
    \item \texttt{tool\_calls} = 8, \texttt{unique\_tools} = 5
    \item \texttt{write\_todos} $\times$2, \texttt{glob} $\times$1, \texttt{temperature\_emissivity\_separation} $\times$2, \texttt{calculate\_band\_mean\_by\_condition} $\times$2, \texttt{difference} $\times$1
\end{itemize}

\textbf{Trajectory (core):} \texttt{temperature\_emissivity\_separation} $\times$2 $\rightarrow$ \texttt{calculate\_band\_mean\_by\_condition} $\times$2 $\rightarrow$ \texttt{difference} $\times$1 (plus \texttt{glob} $\times$1 for discovery)

\textbf{TTM Call Detail:} \texttt{tir\_band\_paths} = [BT\_10, BT\_11, BT\_12] for Polygon1/Polygon2; outputs saved as LST\_TTM\_Polygon1\_calc.tif and LST\_TTM\_Polygon2\_calc.tif

\textbf{Mean/$\Delta$ Detail:} \texttt{calculate\_band\_mean\_by\_condition} with condition \texttt{band0 > 0} produced 282.9996\,K (P1) and 283.7492\,K (P2); difference = 0.7496\,K

\textbf{Key Numbers:}
\begin{center}
\begin{tabular}{ccc}
\toprule
mean(P1) & mean(P2) & $\Delta$LST \\
\midrule
283.00\,K & 283.75\,K & $\approx$0.75\,K \\
\bottomrule
\end{tabular}
\end{center}

\textbf{Final Answer:} \texttt{<Answer>A</Answer>} \quad (GT = A, \textcolor{green!60!black}{\textbf{Correct}})

\textbf{Key Memory:} \textcolor{red}{Call \texttt{temperature\_emissivity\_separation}/\texttt{ttm\_lst} with validated, ordered band arrays + correct parameter names (ensure units are Kelvin).}
\end{promptbox}

\subsubsection{memory cases}

\begin{promptbox}{Non-Parametric Memory: Error Attribution Pattern}
\textbf{Title:} Alignment, Unit and Missing-Data Error Attributions for TVDI Workflows

\textbf{Pattern Type:} Error Attribution \quad | \quad \textbf{Source Problem ID:} 5 \quad | \quad \textbf{Success:} False

\textbf{Description:} Common root causes for incorrect/failed TVDI outputs and representative tool error hints to help diagnose issues quickly.

\textbf{Content:} Remote-sensing guidance that spells out the ordered tool-chain (\texttt{ls} $\rightarrow$ \texttt{verify\_pairs} $\rightarrow$ QA/cloud mask $\rightarrow$ reproject/resample $\rightarrow$ unit conversion $\rightarrow$ \texttt{compute\_tvdi}) with a focus on where failures originate (misalignment, units, masking, nodata propagation).

\textbf{Action Items:}
\begin{enumerate}[nosep,leftmargin=*]
    \item On listing files, immediately check CRS/transform/shape and pixel sizes; if mismatched, reproject/resample before any array arithmetic.
    \item Inspect and apply QA/cloud masks; remove or set nodata consistently so \texttt{compute\_tvdi} does not operate on clouded pixels.
    \item Verify and apply scale and offsets or convert LST units so both inputs are physically compatible prior to TVDI calculation.
    \item Run a small-sample \texttt{compute\_tvdi} on a clipped subset to validate shapes and successful arithmetic before full-batch runs.
    \item If runtime warnings or errors occur, capture and inspect error messages and intermediate arrays (shapes, dtype, NaN counts).
\end{enumerate}

\textbf{Detection Cues:}
\begin{itemize}[nosep,leftmargin=*]
    \item Compute step fails with shape/array broadcasting errors or produces uniform NaN outputs after masking.
    \item Warnings such as ``invalid value encountered'' or arithmetic division warnings during TVDI computation.
    \item Sudden extreme hotspot proportions (very near 0\% or very near 100\%) inconsistent with expectations---suggests masking/units issues.
    \item Output rasters all nodata or many pixels flagged nodata after reprojection/resampling---check nodata propagation and align extents.
\end{itemize}

\textbf{Failure Cause:} \\
\textit{Root causes:} Mismatched spatial alignment (CRS/transform/shape) or resolution leading to broadcast/shape errors; inconsistent LST units or missing scale factors causing incorrect TVDI ranges; inadequate QA/cloud masking producing inflated or NaN results.

\textit{Tool error hints:} Watch for messages like \texttt{"ValueError: operands could not be broadcast together"} or \texttt{"RuntimeWarning: invalid value encountered in true\_divide"} or diagnostics stating \texttt{"No overlapping valid data after masking"}---these point to alignment/masking/unit problems.
\end{promptbox}

\begin{promptbox}{Non-Parametric Memory: Analysis Pattern}
\textbf{Title:} Single-Channel LST from Landsat-8 --- Ordered Tool Chain

\textbf{Pattern Type:} Analysis Pattern \quad | \quad \textbf{Source Problem ID:} 7 \quad | \quad \textbf{Success:} True

\textbf{Description:} Reusable analysis pattern for deriving NDVI and single-channel LST time series from Landsat TOA/thermal scenes with per-image hotspot aggregation and seasonal counting.

\textbf{Content:} Remote-sensing guidance that spells out the ordered tool-chain (Tool A $\rightarrow$ Tool B $\rightarrow$ Tool C) and key decision checkpoints.

\textbf{Action Items:}
\begin{enumerate}[nosep,leftmargin=*]
    \item Call \texttt{ls} or \texttt{glob} with an explicit path argument to list TIFFs and enumerate available bands (thermal band, Red, NIR).
    \item Group files into per-acquisition triplets (BT10, B4, B5) by date; verify filenames and timestamps before batching.
    \item Call \texttt{reproject}/\texttt{resample} as needed to align CRS and pixel size among BT, Red and NIR rasters (ensure identical extent, resolution, and origin).
    \item Call \texttt{cloud\_masking} (QA bitmask or external mask) to mask clouds/shadows and exclude nodata pixels from calculations.
    \item Call \texttt{calculate\_batch\_ndvi} with explicit \texttt{input\_red\_paths} and \texttt{input\_nir\_paths} to produce NDVI rasters (one output per date).
    \item Compute emissivity per scene from NDVI (NDVI threshold/vegetation fraction method) and ensure emissivity units are fractional (0--1).
    \item Call \texttt{lst\_single\_channel} per triplet providing \texttt{bt\_path}, \texttt{red\_path}, \texttt{nir\_path} and a validated \texttt{output\_path} to generate LST rasters (ensure BT units required by routine).
    \item Call \texttt{calc\_batch\_image\_hotspot\_percentage} or equivalent to compute per-image fraction of pixels above the temperature threshold (parameterize threshold, e.g., 315\,K).
    \item Temporally filter results to the target season/window, then aggregate per-day or per-scene counts (apply area weighting if scenes have different valid pixel counts).
    \item Validate outputs by spot-checking LST distributions, histogram bounds, geolocation alignment with reflectance bands, and by comparing a subset to independent ground/other-sensor data if available.
\end{enumerate}

\textbf{Detection Cues:}
\begin{itemize}[nosep,leftmargin=*]
    \item Mismatched CRS or pixel grid warnings when aligning bands (signals to run reprojection/resampling).
    \item High fraction of NaNs or zero counts after masking (indicates missing/invalid masks or misaligned extents).
    \item NDVI values outside expected range ($<-1$ or $>1$) or constant NDVI across scene (indicates wrong input bands or scaling).
    \item LST rasters with impossible values (e.g., extremely low/high temperatures or identical values across image) $\rightarrow$ check unit conversions and BT calculation.
    \item Sudden jumps in per-image hotspot fraction that coincide with missing cloud masking or unmasked striping.
\end{itemize}

\textbf{Failure Cause:} None (successful pattern).
\end{promptbox}

\subsection{raw trajactory}
\begin{promptbox}{Trajectory Q157 - Part 1/5: Initialization}
\small
\textbf{Question:} Based on the sur\_refl\_b01 data of Lake Urmia from January 1 to December 30, 2022, calculate water turbidity over time, determine the turbidity per unit area for each date, and identify the time when the water body experienced the highest level of pollution.

\textbf{Choices:} A. Dec 24: 8497.79 \quad B. Jan 16: 8233.77 \quad C. May 1: 7753.23 \quad D. Mar 2: 5961.99 \quad E. Jul 15: 3030.38

\texttt{[Step 1-2] write\_todos} $\rightarrow$ \texttt{tool\_result}
\begin{verbatim}
Updated todo list to [
  {'content': 'Enumerate files in benchmark/data/question157 to find 
   sur_refl_b01 images', 'status': 'in_progress'},
  {'content': 'Compute turbidity NTU for each sur_refl_b01 file and 
   save outputs to question157/', 'status': 'pending'},
  {'content': 'Compute sum of turbidity and water pixel area for each 
   output; calculate turbidity per unit area', 'status': 'pending'},
  {'content': 'Identify date with highest unit-area turbidity and 
   select matching choice', 'status': 'pending'}]
\end{verbatim}

\texttt{[Step 3-4] ls} $\rightarrow$ \texttt{tool\_result}
\begin{verbatim}
["/benchmark/data/question157/lakeUrmia_sur_refl_b01.tif_2022-01-01.tif",
 "/benchmark/data/question157/lakeUrmia_sur_refl_b01.tif_2022-01-16.tif",
 "/benchmark/data/question157/lakeUrmia_sur_refl_b01.tif_2022-01-31.tif",
 "/benchmark/data/question157/lakeUrmia_sur_refl_b01.tif_2022-02-15.tif",
 "/benchmark/data/question157/lakeUrmia_sur_refl_b01.tif_2022-03-02.tif",
 "/benchmark/data/question157/lakeUrmia_sur_refl_b01.tif_2022-03-17.tif",
 "/benchmark/data/question157/lakeUrmia_sur_refl_b01.tif_2022-04-01.tif",
 "/benchmark/data/question157/lakeUrmia_sur_refl_b01.tif_2022-04-16.tif",
 "/benchmark/data/question157/lakeUrmia_sur_refl_b01.tif_2022-05-01.tif",
 "/benchmark/data/question157/lakeUrmia_sur_refl_b01.tif_2022-05-16.tif",
 "/benchmark/data/question157/lakeUrmia_sur_refl_b01.tif_2022-05-31.tif",
 "/benchmark/data/question157/lakeUrmia_sur_refl_b01.tif_2022-06-15.tif",
 "/benchmark/data/question157/lakeUrmia_sur_refl_b01.tif_2022-06-30.tif",
 "/benchmark/data/question157/lakeUrmia_sur_refl_b01.tif_2022-07-15.tif",
 "/benchmark/data/question157/lakeUrmia_sur_refl_b01.tif_2022-07-30.tif",
 "/benchmark/data/question157/lakeUrmia_sur_refl_b01.tif_2022-08-14.tif",
 "/benchmark/data/question157/lakeUrmia_sur_refl_b01.tif_2022-08-29.tif",
 "/benchmark/data/question157/lakeUrmia_sur_refl_b01.tif_2022-09-13.tif",
 "/benchmark/data/question157/lakeUrmia_sur_refl_b01.tif_2022-09-28.tif",
 "/benchmark/data/question157/lakeUrmia_sur_refl_b01.tif_2022-10-25.tif",
 "/benchmark/data/question157/lakeUrmia_sur_refl_b01.tif_2022-11-09.tif",
 "/benchmark/data/question157/lakeUrmia_sur_refl_b01.tif_2022-11-24.tif",
 "/benchmark/data/question157/lakeUrmia_sur_refl_b01.tif_2022-12-09.tif",
 "/benchmark/data/question157/lakeUrmia_sur_refl_b01.tif_2022-12-24.tif"]
\end{verbatim}

\texttt{[Step 5-6] write\_todos} $\rightarrow$ \texttt{tool\_result}
\begin{verbatim}
Updated todo list to [...'Enumerate files'...'status': 'completed'}, 
  {'content': 'Compute turbidity NTU...', 'status': 'in_progress'}...]
\end{verbatim}
\end{promptbox}

\begin{promptbox}{Trajectory Q157 - Part 2/5: Turbidity Batch 1}
\small
\texttt{[Step 7] calculate\_water\_turbidity\_ntu} (tool\_call)\\
\texttt{[Step 8] calculate\_water\_turbidity\_ntu} (tool\_call)\\
\texttt{[Step 9] calculate\_water\_turbidity\_ntu} (tool\_call)\\
\texttt{[Step 10] calculate\_water\_turbidity\_ntu} (tool\_call)\\
\texttt{[Step 11] calculate\_water\_turbidity\_ntu} (tool\_call)\\
\texttt{[Step 12] calculate\_water\_turbidity\_ntu} (tool\_call)\\
\texttt{[Step 13] calculate\_water\_turbidity\_ntu} (tool\_call)\\
\texttt{[Step 14] calculate\_water\_turbidity\_ntu} (tool\_call)\\
\texttt{[Step 15] calculate\_water\_turbidity\_ntu} (tool\_call)\\
\texttt{[Step 16] calculate\_water\_turbidity\_ntu} (tool\_call)\\
\texttt{[Step 17] calculate\_water\_turbidity\_ntu} (tool\_call)\\
\texttt{[Step 18] calculate\_water\_turbidity\_ntu} (tool\_call)

\texttt{[Step 19] tool\_result:} \verb|Result saved at .../turbidity_2022-01-01.tif|\\
\texttt{[Step 20] tool\_result:} \verb|Result saved at .../turbidity_2022-01-16.tif|\\
\texttt{[Step 21] tool\_result:} \verb|Result saved at .../turbidity_2022-01-31.tif|\\
\texttt{[Step 22] tool\_result:} \verb|Result saved at .../turbidity_2022-02-15.tif|\\
\texttt{[Step 23] tool\_result:} \verb|Result saved at .../turbidity_2022-03-02.tif|\\
\texttt{[Step 24] tool\_result:} \verb|Result saved at .../turbidity_2022-03-17.tif|\\
\texttt{[Step 25] tool\_result:} \verb|Result saved at .../turbidity_2022-04-01.tif|\\
\texttt{[Step 26] tool\_result:} \verb|Result saved at .../turbidity_2022-04-16.tif|\\
\texttt{[Step 27] tool\_result:} \verb|Result saved at .../turbidity_2022-05-01.tif|\\
\texttt{[Step 28] tool\_result:} \verb|Result saved at .../turbidity_2022-05-16.tif|\\
\texttt{[Step 29] tool\_result:} \verb|Result saved at .../turbidity_2022-05-31.tif|\\
\texttt{[Step 30] tool\_result:} \verb|Result saved at .../turbidity_2022-06-15.tif|

\texttt{[Step 31-32] write\_todos} $\rightarrow$ \texttt{tool\_result} (update progress)
\end{promptbox}

\begin{promptbox}{Trajectory Q157 - Part 3/5: Turbidity Batch 2}
\small
\texttt{[Step 33] calculate\_water\_turbidity\_ntu} (tool\_call)\\
\texttt{[Step 34] calculate\_water\_turbidity\_ntu} (tool\_call)\\
\texttt{[Step 35] calculate\_water\_turbidity\_ntu} (tool\_call)\\
\texttt{[Step 36] calculate\_water\_turbidity\_ntu} (tool\_call)\\
\texttt{[Step 37] calculate\_water\_turbidity\_ntu} (tool\_call)\\
\texttt{[Step 38] calculate\_water\_turbidity\_ntu} (tool\_call)\\
\texttt{[Step 39] calculate\_water\_turbidity\_ntu} (tool\_call)\\
\texttt{[Step 40] calculate\_water\_turbidity\_ntu} (tool\_call)\\
\texttt{[Step 41] calculate\_water\_turbidity\_ntu} (tool\_call)\\
\texttt{[Step 42] calculate\_water\_turbidity\_ntu} (tool\_call)\\
\texttt{[Step 43] calculate\_water\_turbidity\_ntu} (tool\_call)\\
\texttt{[Step 44] calculate\_water\_turbidity\_ntu} (tool\_call)

\texttt{[Step 45] tool\_result:} \verb|Result saved at .../turbidity_2022-06-30.tif|\\
\texttt{[Step 46] tool\_result:} \verb|Result saved at .../turbidity_2022-07-15.tif|\\
\texttt{[Step 47] tool\_result:} \verb|Result saved at .../turbidity_2022-07-30.tif|\\
\texttt{[Step 48] tool\_result:} \verb|Result saved at .../turbidity_2022-08-14.tif|\\
\texttt{[Step 49] tool\_result:} \verb|Result saved at .../turbidity_2022-08-29.tif|\\
\texttt{[Step 50] tool\_result:} \verb|Result saved at .../turbidity_2022-09-13.tif|\\
\texttt{[Step 51] tool\_result:} \verb|Result saved at .../turbidity_2022-09-28.tif|\\
\texttt{[Step 52] tool\_result:} \verb|Result saved at .../turbidity_2022-10-25.tif|\\
\texttt{[Step 53] tool\_result:} \verb|Result saved at .../turbidity_2022-11-09.tif|\\
\texttt{[Step 54] tool\_result:} \verb|Result saved at .../turbidity_2022-11-24.tif|\\
\texttt{[Step 55] tool\_result:} \verb|Result saved at .../turbidity_2022-12-09.tif|\\
\texttt{[Step 56] tool\_result:} \verb|Result saved at .../turbidity_2022-12-24.tif|

\texttt{[Step 57-58] write\_todos} $\rightarrow$ \texttt{tool\_result} (mark completed)
\end{promptbox}

\begin{promptbox}{Trajectory Q157 - Part 4/5: Sum \& Area Calculation}
\small
\texttt{[Step 59] calc\_batch\_image\_sum} (tool\_call)\\
\texttt{[Step 60] tool\_result:}
\begin{verbatim}
[392065568.0, 416365280.0, 219604864.0, 129923824.0, 301486400.0, 
 238540544.0, 217505424.0, 104889288.0, 292429280.0, 148286080.0, 
 112012264.0, 127280272.0, 123388280.0, 153240192.0, 146451488.0, 
 166537264.0, 155292112.0, 170321376.0, 154315408.0, 135877632.0, 
 183978176.0, 152578496.0, 125695640.0, 429716032.0]
\end{verbatim}

\texttt{[Step 61-72] calculate\_area} (12 parallel tool\_calls)\\
\texttt{[Step 73] tool\_result:} \verb|50568.0|\\
\texttt{[Step 74] tool\_result:} \verb|50568.0|\\
\texttt{[Step 75] tool\_result:} \verb|50568.0|\\
\texttt{[Step 76] tool\_result:} \verb|50553.0|\\
\texttt{[Step 77] tool\_result:} \verb|50557.0|\\
\texttt{[Step 78] tool\_result:} \verb|50568.0|\\
\texttt{[Step 79] tool\_result:} \verb|50568.0|\\
\texttt{[Step 80] tool\_result:} \verb|50476.0|\\
\texttt{[Step 81] tool\_result:} \verb|50568.0|\\
\texttt{[Step 82] tool\_result:} \verb|50568.0|\\
\texttt{[Step 83] tool\_result:} \verb|50547.0|\\
\texttt{[Step 84] tool\_result:} \verb|50512.0|

\texttt{[Step 85-96] calculate\_area} (12 parallel tool\_calls)\\
\texttt{[Step 97] tool\_result:} \verb|50526.0|\\
\texttt{[Step 98] tool\_result:} \verb|50568.0|\\
\texttt{[Step 99] tool\_result:} \verb|50568.0|\\
\texttt{[Step 100] tool\_result:} \verb|50568.0|\\
\texttt{[Step 101] tool\_result:} \verb|50559.0|\\
\texttt{[Step 102] tool\_result:} \verb|50568.0|\\
\texttt{[Step 103] tool\_result:} \verb|50560.0|\\
\texttt{[Step 104] tool\_result:} \verb|50568.0|\\
\texttt{[Step 105] tool\_result:} \verb|50568.0|\\
\texttt{[Step 106] tool\_result:} \verb|50568.0|\\
\texttt{[Step 107] tool\_result:} \verb|50568.0|\\
\texttt{[Step 108] tool\_result:} \verb|50568.0|
\end{promptbox}

\begin{promptbox}{Trajectory Q157 - Part 5/5: Final Computation}
\small
\texttt{[Step 109-110] write\_todos} $\rightarrow$ \texttt{tool\_result}
\begin{verbatim}
Updated todo list to [
  {'content': 'Enumerate files...', 'status': 'completed'},
  {'content': 'Compute turbidity NTU...', 'status': 'completed'},
  {'content': 'Compute sum and area...', 'status': 'completed'},
  {'content': 'Identify date with highest...', 'status': 'in_progress'}]
\end{verbatim}

\texttt{[Step 111] diagnostics}
\begin{verbatim}
Tool summary:
- write_todos: Created and updated 4-step todo list
- ls: Listed 24 input sur_refl_b01 files (2022-01-01 to 2022-12-24)
- calculate_water_turbidity_ntu (batch 1, 12 calls): 2022-01-01...06-15
- calculate_water_turbidity_ntu (batch 2, 12 calls): 2022-06-30...12-24
- calc_batch_image_sum: Computed sums for all 24 turbidity images
- calculate_area (batch 1, 12 calls): Computed pixel counts
- calculate_area (batch 2, 12 calls): Computed pixel counts
- Local computation: turbidity_per_unit = sum / area

Failure reason: None
\end{verbatim}

\textbf{Final Computation:}
\begin{verbatim}
2022-12-24: 429716032.0 / 50568.0 = 8497.79  <-- MAXIMUM
2022-01-16: 416365280.0 / 50568.0 = 8233.77
2022-03-02: 301486400.0 / 50557.0 = 5961.99
2022-07-15: 153240192.0 / 50568.0 = 3030.38
\end{verbatim}

\textbf{Output:} \texttt{<Answer>A</Answer>}

\end{promptbox}

\newpage
\subsection{Failure Analysis}
\label{app:fail}
\begin{figure}[h]
  \centering
  \includegraphics[width=0.8\columnwidth]{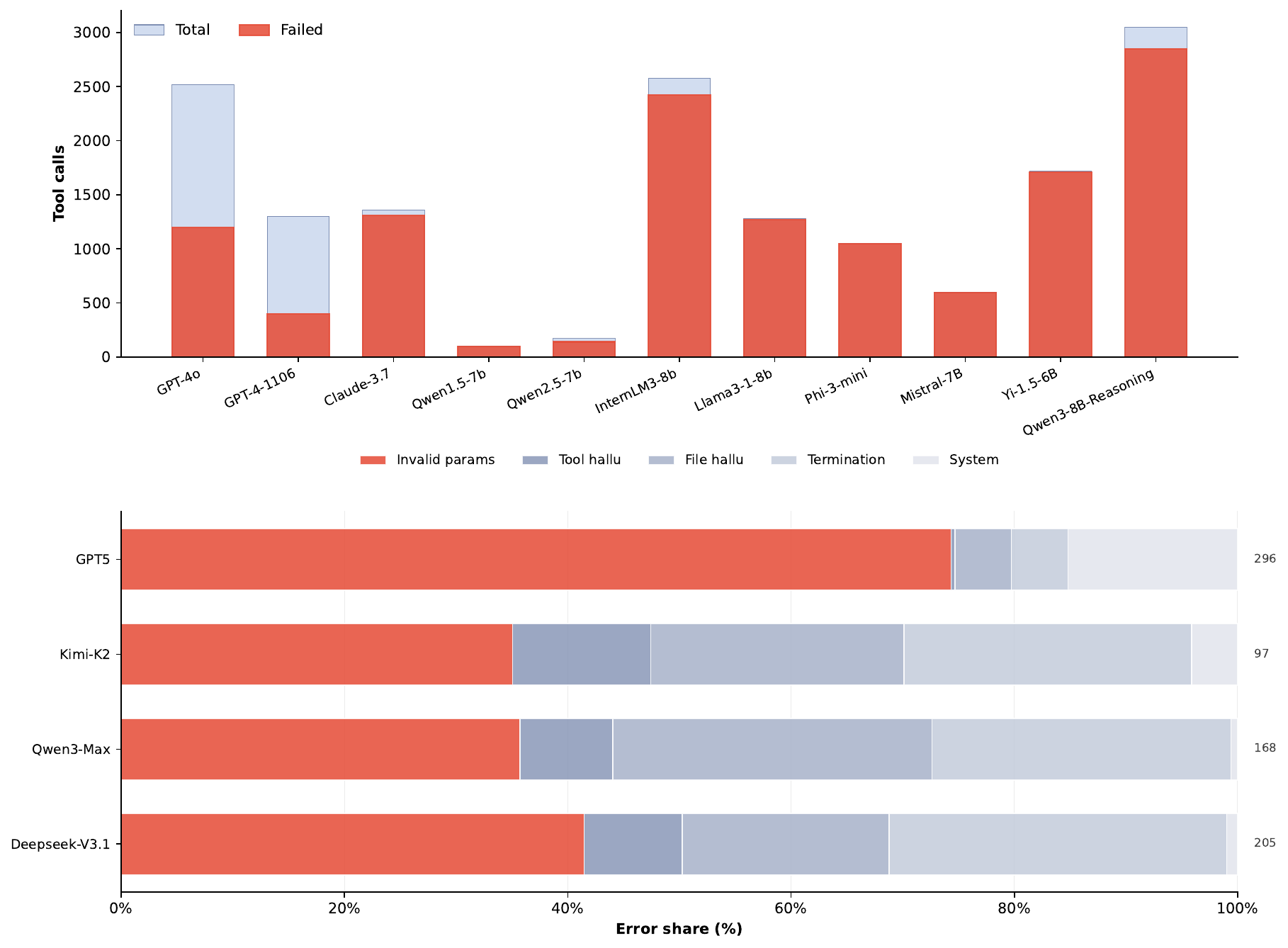}
  \caption{Fail reasons from previous earth agent benchmarks~\cite{shabbir2025thinkgeo,feng2025earth}}
  \label{fig:fail_reason}
\end{figure}

\begin{figure}[h]
  \centering
  \includegraphics[width=0.8\columnwidth]{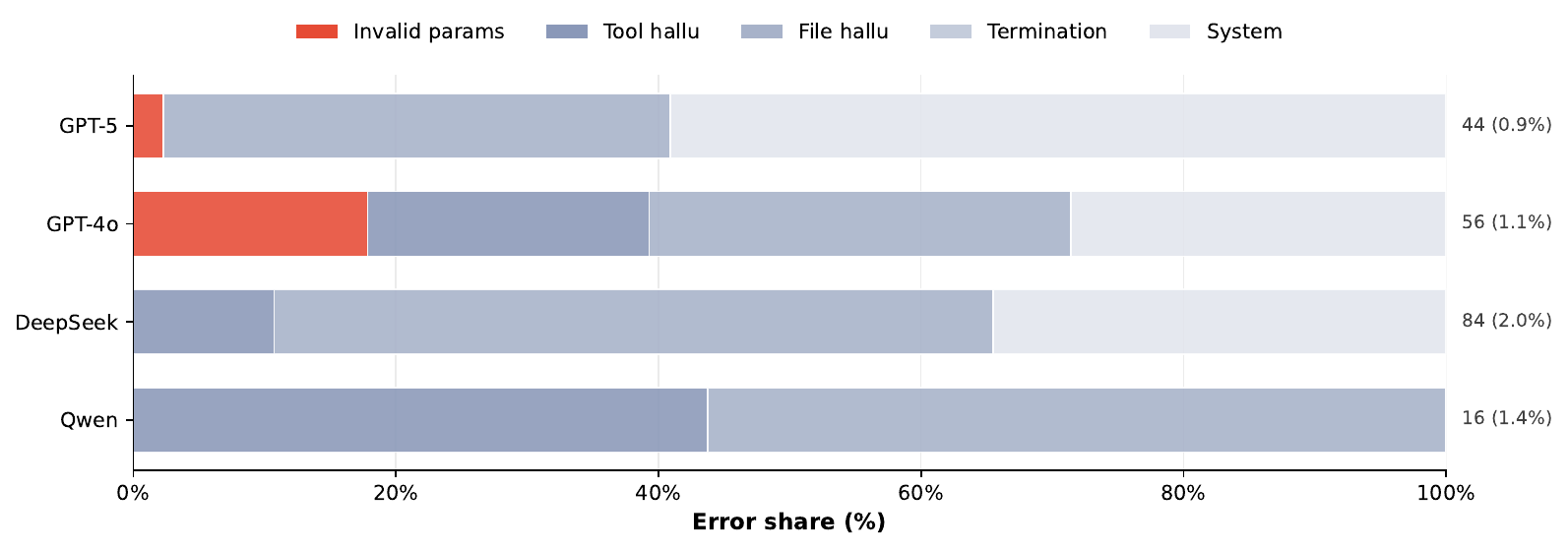}
  \caption{Fail reasons for GeoEvolver on Earth Agent Benchmark}
  \label{fig:fail_reason}
\end{figure}

\end{document}